\documentclass[11pt]{article}

\usepackage[]{ACL2023}
\usepackage{graphicx}
\usepackage{caption}
\usepackage{subcaption}
\usepackage{tabularx}
\usepackage{amsmath}
\usepackage{amssymb}
\usepackage{booktabs}
\usepackage{pbox}

\usepackage{xcolor}

\newcommand{\eg}{\emph{e.g.}}

\newcommand{\method}{Parallel Context Windows}
\newcommand{\methodshort}{PCW}
\newcommand{\icl}{in-context learning}
\newcommand{\iclshort}{ICL}

\usepackage{times}
\usepackage{latexsym}

\usepackage{tcolorbox}

\usepackage[T1]{fontenc}

\usepackage[utf8]{inputenc}

\usepackage{microtype}

\usepackage{inconsolata}

\title{Parallel Context Windows for Large Language Models}

\author{Nir Ratner~~~~Yoav Levine~~~~Yonatan Belinkov~~~~Ori Ram~~~~Inbal Magar~~~~Omri Abend\\
\textbf{ Ehud Karpas~~~Amnon Shashua~~~Kevin Leyton-Brown~~~Yoav Shoham}\\
  AI21 Labs\\
  \texttt{nirr@ai21.com} }
\begin{document}
\maketitle
\begin{abstract}
When applied to processing long text, Large Language Models (LLMs) are limited by their context window.
Existing efforts to address this  limitation involve training specialized architectures, and cannot be easily applied to off-the-shelf LLMs.
We present Parallel Context Windows (PCW), a method that alleviates the context window restriction for any off-the-shelf LLM \textit{without further training}. 
The key to the approach is to carve a long context into chunks (``windows''), restrict the attention mechanism to apply only within each window, and re-use the positional embeddings across the windows. 
Our main results test the PCW approach on in-context learning with models that range in size between 750 million and 178 billion parameters, and show substantial improvements for tasks with diverse input and output spaces. We show additional benefits in other settings where long context windows may be beneficial: multi-hop questions and retrieval-augmented question answering with multiple retrieved documents.
Our results highlight Parallel Context Windows as a promising method for applying off-the-shelf LLMs in a range of settings that require long text sequences.~{We make our code publicly available at \url{https://github.com/ai21labs/parallel-context-windows}.}
\end{abstract}

\section{Introduction}

 \begin{figure}[t!]
    \includegraphics[clip,width=\columnwidth]{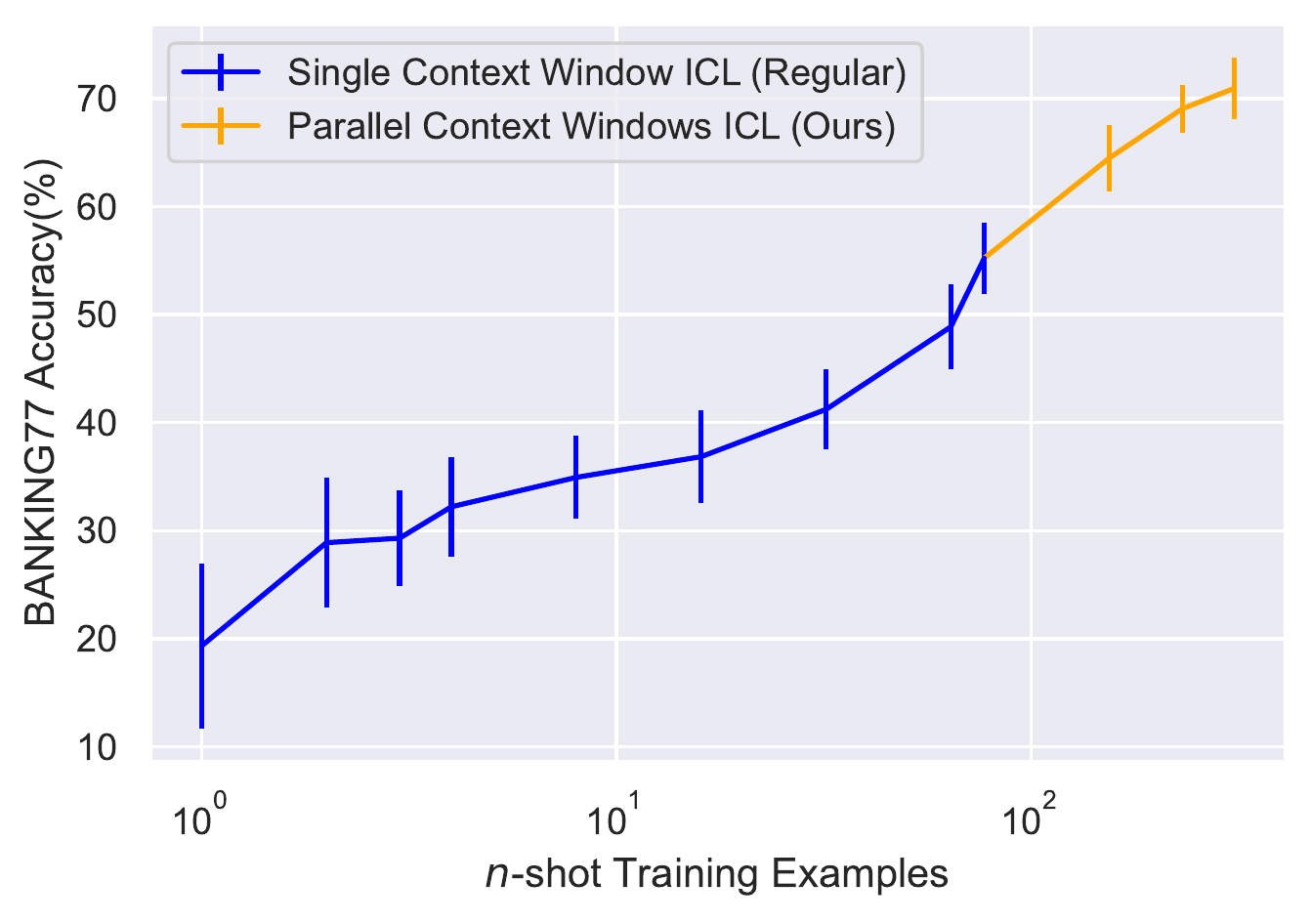}
     \centering
        \caption{In-context learning (\iclshort{}) accuracy against $n$-shot training examples for the BANKING77 intent classification dataset \cite{banking77} using the model Jurassic-1-Grande (17B). The blue line shows the improvement in performance as the context window is filled with examples; the orange line shows how our \method{} method, which adds up to four times more training examples, provides a significant boost in performance. The error bars represent the standard deviation across multiple runs, as explained in Section \ref{sec:setup}.}
        \label{fig:Grande banking77}
\end{figure}

\begin{figure*}[t!]
    \vspace{-20pt}
    \includegraphics[clip,width=0.99\textwidth]{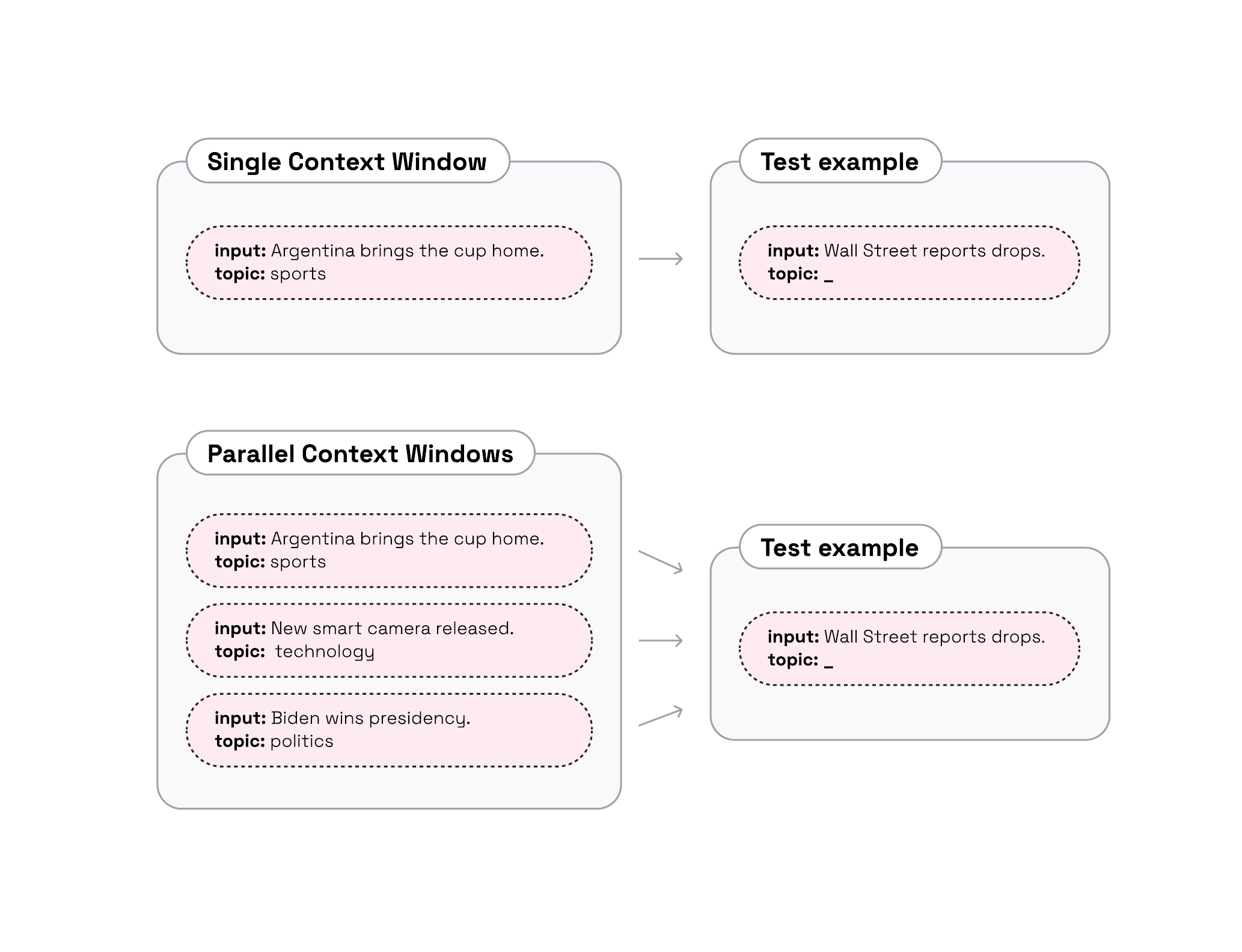}
     \centering
     \vspace{-30pt}
        \caption{An illustration of Parallel Context Windows (PCW) approach, exposing the LLM to text within multiple context windows during generation. Tokens inside each window attend only to the previous tokens \textit{in their window}. Test example tokens attend to the tokens of all context windows.}
        \label{fig:illustration}
\end{figure*}

A key parameter of a Large Language Model (LLM) is its \emph{context window}, the number of text tokens  it can process in a forward pass. 
Current LLM architectures limit the context window size---typically up to several thousand tokens---
because the global nature of the attention mechanism imposes computational costs quadratic in context length.
This presents an obstacle to use cases where the LLM needs to process a lot of text, \eg, tackling tasks that require long inputs~\cite{tay2020long,shaham2022scrolls}, considering large sets of retrieved documents for open-book question answering~\cite{karpukhin2020dense,levine2022standing,levine2022huge}, or performing in-context learning~\cite{gpt3} when the desired input-output relationship cannot be adequately characterized within the context window.

Previous work has addressed such obstacles by training dedicated architectures, \eg, training sparse attention mechanisms for long inputs~\cite{bigbird,longT5} and Fusion-in-Decoder readers for retrieved documents~\cite{izacard2020leveraging}. 
However, these architectures are often tailored to specific use cases, and they are often constrained in terms of their size as a trade-off, in order to facilitate long text consumption.
It remains an open problem to find an effective way to allow off-the-shelf LLMs to process text longer than its original context window,
\textit{without dedicated training}. 

In this paper, we introduce \textit{Parallel Context Windows} (PCW), illustrated in Figure \ref{fig:illustration}, a 
new approach for addressing this problem in any decoder-based LLM\footnote{We will use LLM to refer to decoder-only language models.}, and show its efficacy in several setups.  
PCW involves splitting long text into multiple parallel contexts, each equally accessible during output generation. 
Doing so consists of two simple \textit{post-hoc} modifications to a pretrained LLM, neither of which requires any further training:
(1) using sparse masking to allow each context window to attend only to itself, while still allowing the generated text to attend to all contexts simultaneously; and
(2) reusing the model's positional embeddings within each parallel context window, 
sidestepping the problem of extrapolating positional embeddings and signaling to the model that each window is equally ``close'' to the generated tokens.

We conducted an in-depth investigation of the extent to which Parallel Context Windows can improve LLMs' ability to perform \textit{in-context learning}
~\cite{gpt3}: when a pretrained LLM is given an input sequence of concatenated ``training'' input--output pairs representing a task, 
followed by a single ``test'' input, it is able to supply the corresponding test output with high accuracy. 
Crucially, in the setting of in-context learning, the context window limitation inherently caps the number of training examples that can be inserted before the test example.
This significantly limits the applicability of in-context learning for tasks with long or highly diverse inputs or outputs.

We focus on these types of tasks, showing that Parallel Context Windows significantly aid in-context learning of two task families that tend to suffer from low in-context learning performance: classification tasks that have many classes and extractive question answering tasks. 
We experiment with GPT2 models \citep{gpt2} having between 750M and 1.5B parameters, LLaMA models \citep{llama} having between 7B and 65B parameters, and Jurassic-1 models \citep{lieber2021jurassic} having between 7B and 178B parameters. 
Notably, using $3$ Parallel Context Windows for classification tasks with more than $6$ classes results in average performance gains of $6.7$ and $7.1$ points for LLaMA models 32.5B and 65B, respectively, and $7.4$, $8.2$, and $8.7$ gains for Jurassic-1 models 7B, 17B, and 178B, respectively. (see example in Figure~\ref{fig:Grande banking77}).
Our results show that Parallel Context Windows  broadens the scope of tasks that can be learned via the popular setup of in-context learning, to tasks that require more training examples than permitted in current context sizes.

We further explore the applicability of PCW to two other settings that may benefit from the integration of several documents. One is multi-hop question answering, where the different pieces of information are shown in different windows. 
We show that in some cases parallel reading is beneficial, through a test case on the HotpotQA benchmark \citep{yang-etal-2018-hotpotqa}. The other setting is retrieval-augmented question answering, where we show that reading several retrieved documents in parallel is advantageous, through a test case on the Natural Questions benchmark \citep{kwiatkowski-etal-2019-natural}.

Overall, we provide clear evidence that, without any further training, Parallel Context Windows can make a large amount of text accessible to an off-the-shelf LLM during decoding. We thus see promise in further investigation of  Parallel Context Windows for applying off-the-shelf LLMs in other applications that require such capabilities, such as tackling tasks with long inputs.

 \begin{figure}[t!]
    \includegraphics[clip,width=\columnwidth]{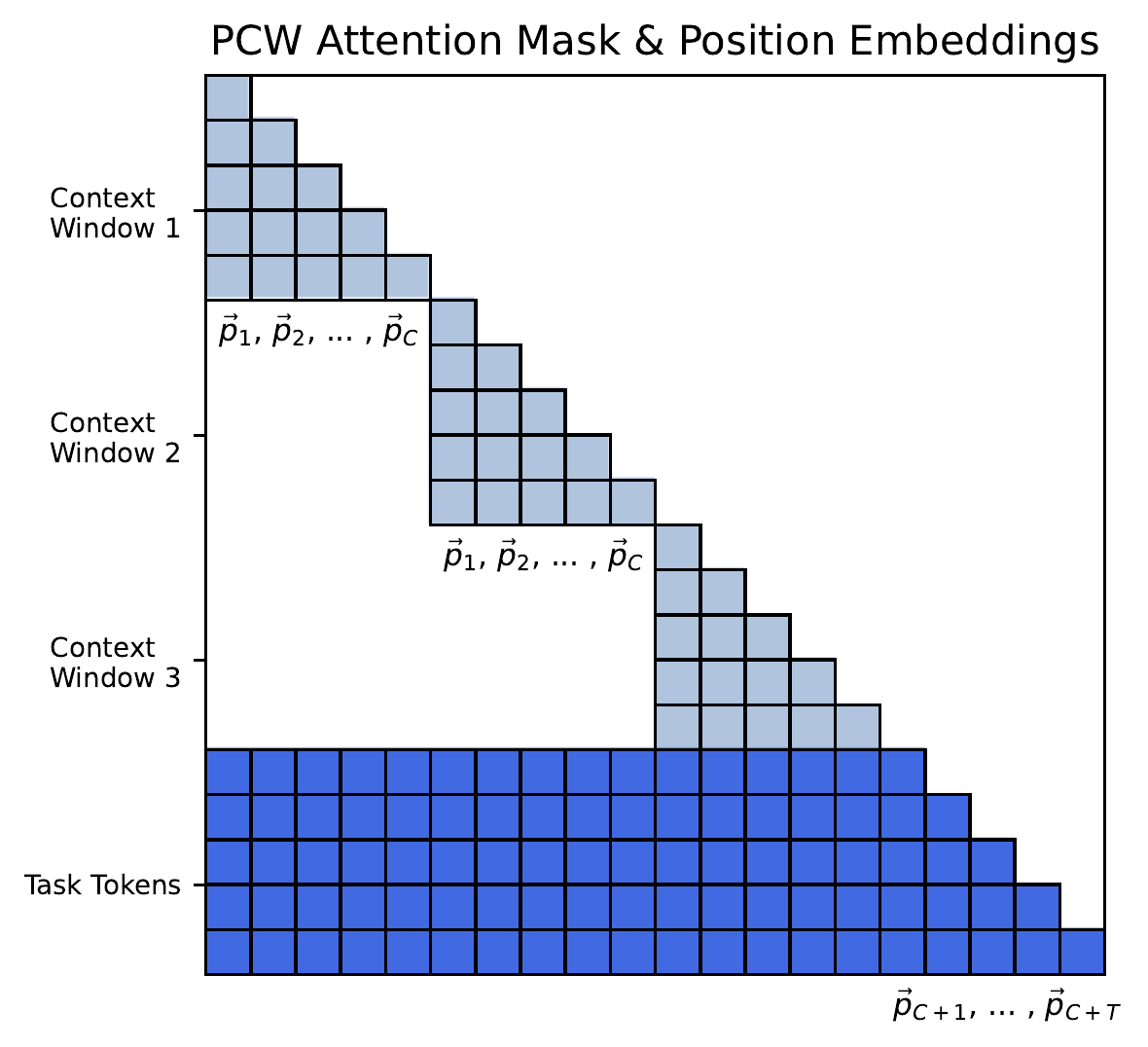}
     \centering
        \caption{
        Attention pattern and positional embeddings assignment in \methodshort{}.
        The $(i,j)$ cell in the matrix is colored iff  the $i^{th}$ token can attend to the $j^{th}$ token.
        Each context window (in grey) attends to itself and is assigned positional embeddings ($\vec{p}_i$) independently, thus re-using the positional vectors. Task tokens (in blue) attend to all the windows. \methodshort{} makes the attention matrix sparser, effectively parallelizing the processing of multiple windows.}
        \label{fig:attention masks}
\end{figure}

\section{Parallel Context Windows} \label{sec:method}

This section provides the details of our \method~method.
The high-level idea of PCW is to insert a long input sequence into multiple replicas of the LLM's original context window, and to allow for a small amount of tokens at the end of the sequence to attend to all of the context windows simultaneously. 
We design \methodshort{} so that the modifications made to the off-the-shelf LLM are minimal, such that processing long contexts remains effective even without further training of the LLM.  
A side advantage is that the LLM modifications required for PCW are quite simple to implement. 
Specifically, PCW applies two modifications to two mechanisms in common autoregressive LLMs: the positional embeddings (Section~\ref{sec:2.1}) and the attention mask (Section~\ref{sec:2.2}).  Figure~\ref{fig:attention masks} illustrates both changes.

\subsection{Positional Embeddings Modification}\label{sec:2.1}

Denoting the LLM's original context window size by $N$ and the Transformer's input representation dimension by $d$, Transformer-based LLMs receive information regarding the input text ordering via a set of $N$ positional embeddings $\{\Vec{p}_i\in\mathbb{R}^d\}_{i=1}^{N}$, by adding $\Vec{p}_i$ to the input token embedding in position $i$\footnote{Simple addition is applied for absolute positional embeddings; for rotary positional embeddings~\cite{roformer} we multiply $\Vec{p}_i$ by the keys and queries in each layer.}.

We conceptually divide the tokens at the input of the LLM into \textit{context tokens} and \textit{task tokens}. The context tokens are inputs that assist the LLM with a given task, such as in-context examples, or relevant retrieved documents. Task tokens refer to the input of the test example, \eg, a sentence to be classified or a question.

When considering a task that requires $T$ task tokens to formulate, the fact that there are only $N$ trained positional embeddings implies that effectively only $C=N-T$ input tokens can be processed as context.\footnote{In the case of absolute positional embeddings this is a hard restriction; for relative positional embeddings, processing more tokens entails degradation~\cite{press2021train}.}
In order to implement \methodshort, we expand the number of processable context tokens by a factor of $B$ such that the overall input sequence can include $B\cdot C + T$ tokens. In order to allow LLMs to process this long sequence of text, we assign one of $N$ positional embedding vectors to location $i\in\{1,\ldots,B\cdot C + T\}$ by the following mapping (depicted in Figure~\ref{fig:attention masks}):
\begin{align}
\Vec{p}_i^{PCW} = 
    \begin{cases}
        \Vec{p}_{(i-1~{\textrm{mod}}~C)+1} & {\small 1\leq i \leq B C}\\
        \Vec{p}_{i-(B-1)C} & {\small B C < i \leq B C +T}
    \end{cases}
\end{align}
In words, via this mapping, the model effectively identifies $B$ replicas of the first $C$ original positional embeddings, and the $T$ task tokens retain the last $T$ positional embeddings, now seeing these $B$ 
 replicas as context in their near past. 
 We refer to these replicas of the positional embeddings as \textit{context window replicas}. 
 Notably, while the above re-use of the positional embeddings assigns meaningful positions to all tokens within the longer input sequence, the memory cost of this expansion is quadratic, and moreover, the model was not trained to have two tokens in the same position attend to each other. To address these,  we describe below a modification to the LLM's attention mechanism.

\subsection{Attention Mask Modification}\label{sec:2.2}

We impose a restriction on the attention mechanism which implies that tokens within each context window replica perform autoregressive attention to other tokens in their context window replica, and do not attend to tokens in other context window replicas. In contrast, the task tokens attend to context tokens within all context window replicas.

In the above setting of context window size $N$, we represent attention restrictions by attention mask scores $a_{ii'}\in\{0,1\}$ for $i,i'\in[N]:=\{1,\ldots,N\}$. If $a_{ii'}=0$ then for any Transformer layer in the LLM, tokens in input location $i$ cannot attend to tokens in input location $i'$,  and if $a_{ii'}=1$ they can. 
In common autoregressive LLMs, a token can only attend to tokens that precede it, which following the above notation is translated into $ a_{ii'}=1$ if $1\leq i' \leq i \leq N$ and $ a_{ii'}=0$ otherwise. 

For the case of \methodshort, the $B$ parallel context windows include tokens in positions $i\in[C]$, and are identified with an index $b\in[B]$. The $T$ task tokens are not parallelized, and are located in positions $i\in\{C+1,\ldots,C+T=N\}$. 
For completeness of the notation, we will assign a dummy context window index $b=B+1$ to the $T$ task tokens. 
We add a second index to the attention scores:  $a^{bb'}_{ii'}\in\{0,1\}$ for $i,i'\in[N]$ and $b,b'\in[B]$. 
Similarly to the above, if $a^{b,b'}_{ii'}=0$ then for any Transformer layer in the LLM, tokens in  input location $i$ and context window $b$ cannot attend to tokens in input location $i'$ and context window $b'$, and if $a^{b,b'}_{ii'}=1$ they can. 

With the above notation in place, the following restriction implies that context tokens perform autoregressive attention within each context window replica (illustrated in Figure~\ref{fig:attention masks}):

\begin{equation}
    a^{b,b'}_{ii'}=
    \begin{cases}
    1,  & \text{if } 1\leq i' \leq i \leq C ~~\textrm{and} ~~b=b'\\
    0,  & \text{otherwise}
    \end{cases}
\end{equation}
The following implies that the $T$ task tokens attend to all tokens in all $B$ context windows (for $i>C$):
\begin{equation}
    a^{B+1,b'}_{ii'}=
    \begin{cases}
    1,  & \text{if } 1\leq i' \leq i \leq N,~~ b'\in[B+1]\\
    0,  & \text{otherwise}
    \end{cases}
\end{equation}

The above attention masks allow the model to attend to $B$ times more context when decoding the output, while keeping the computational cost linear in the number of parallel contexts $B$. Overall, for both the above PCW~modifications, assigning $B=1$ corresponds to the vanilla LLM mechanism.

\section{PCW for In-Context Learning}  \label{sec:icl}
\subsection{Experimental Setup} \label{sec:setup}

We apply the \methodshort{} method in the setting of in-context learning (ICL): 
we distribute the in-context training examples among the multiple context window replicas, thus allowing the test example to attend to more training examples.
 For each experiment, we report the performance with regular ICL, using the maximum number of examples that fit in a model's context window ($n_{max}$).  For our \methodshort~method, given $B$ parallel windows, we effectively use $B \times n_{max}$ training examples. The $n_{max}$ used for each dataset and model can be found in Table \ref{tab:prompts}.
Unless stated otherwise, we report results with $B=3$ in the main paper, and discuss the choice of $B$ in Appendix \ref{app:b_values}. 
Since training examples vary in length, we allocate in-context examples into the parallel windows in a manner that balances the windows' lengths.\footnote{Within each window, positional embeddings are assigned sequentially starting from $1$. See Appendix \ref{app:exp-details} for a discussion.}
The test example (corresponding to the $T$ task tokens in Section~\ref{sec:method}) receives the positional embedding that immediately follows the longest context window.

\begin{table*}[t!]
    \centering
\resizebox{0.83\textwidth}{!}{%
\begin{tabular}{lc ll ll  ll }
\toprule
          &     & \multicolumn{2}{c}{J1- Large (7.5B)} & \multicolumn{2}{c}{J1-Grande (17B)} & \multicolumn{2}{c}{J1-Jumbo (178B)} \\
          \cmidrule(lr){3-4} \cmidrule(lr){5-6} \cmidrule(lr){7-8}
         Dataset &  \#  Labels  &                          ICL &               \methodshort{} &                        ICL &               \methodshort{} &                        ICL &               \methodshort{} \\
         \midrule
SST-2 & 2   &                 $93.5_{1.6}$ &    $\boldsymbol{93.8_{1.1}}$ &               $95.2_{1.1}$ &    $\boldsymbol{95.6_{0.5}}$ &               $96.5_{1.4}$ &    $\boldsymbol{97.0_{1.5}}$ \\
CR & 2   &                 $\boldsymbol{93.9_{0.7}}$ &                 $\boldsymbol{93.9_{0.7}}$ &               $93.6_{0.8}$ &    $\boldsymbol{93.8_{0.8}}$ &  $\boldsymbol{93.6_{1.5}}$ &                 $93.1_{1.0}$ \\
RTE & 2   &    $\boldsymbol{58.3_{3.8}}$ &                 $58.1_{3.7}$ &               $61.2_{5.1}$ &    $\boldsymbol{62.2_{3.0}}$ &               $63.9_{5.0}$ &    $\boldsymbol{66.0_{4.1}}$ \\
Subj & 2   &  $\boldsymbol{84.1_{7.7}}^*$ &                 $79.1_{7.2}$ &               $93.0_{2.5}$ &  $\boldsymbol{95.3_{1.2}}^*$ &               $89.1_{5.3}$ &  $\boldsymbol{93.6_{2.1}}^*$ \\
CB & 3   &    $\boldsymbol{65.2_{8.0}}$ &                 $61.2_{8.2}$ &               $75.0_{8.1}$ &    $\boldsymbol{75.7_{6.0}}$ &               $76.2_{4.3}$ &    $\boldsymbol{76.6_{3.5}}$ \\
AGNews & 4   &                 $79.8_{3.6}$ &  $\boldsymbol{81.5_{2.1}}^*$ &               $81.4_{3.0}$ &    $\boldsymbol{82.7_{2.1}}$ &               $82.5_{3.8}$ &  $\boldsymbol{85.9_{1.7}}^*$ \\
SST-5 & 5   &                 $45.5_{3.9}$ &  $\boldsymbol{47.4_{2.9}}^*$ &               $51.6_{3.4}$ &  $\boldsymbol{53.8_{2.2}}^*$ &  $\boldsymbol{55.4_{2.8}}$ &                 $55.1_{3.9}$ \\
YELP & 5   &                 $56.2_{3.8}$ &    $\boldsymbol{56.3_{5.1}}$ &  $\boldsymbol{66.2_{2.2}}$ &                 $65.6_{2.0}$ &  $\boldsymbol{66.3_{4.1}}$ &                 $65.4_{2.6}$ \\
TREC & 6   &                 $87.0_{4.5}$ &  $\boldsymbol{89.4_{3.2}}^*$ &               $86.5_{3.8}$ &  $\boldsymbol{88.7_{3.4}}^*$ &               $87.1_{5.7}$ &    $\boldsymbol{90.4_{3.1}}$ \\
DBPedia & 14  &                 $93.2_{3.0}$ &  $\boldsymbol{96.2_{1.5}}^*$ &               $92.5_{3.3}$ &  $\boldsymbol{97.3_{1.6}}^*$ &               $91.7_{4.4}$ &  $\boldsymbol{96.5_{2.3}}^*$ \\
NLU Scenario & 18  &                 $81.9_{2.2}$ &  $\boldsymbol{84.2_{1.5}}^*$ &               $86.1_{2.1}$ &  $\boldsymbol{88.8_{1.1}}^*$ &               $85.4_{2.9}$ &  $\boldsymbol{87.8_{1.6}}^*$ \\
TREC Fine & 50  &                 $60.5_{6.9}$ &  $\boldsymbol{68.8_{3.4}}^*$ &               $63.3_{6.0}$ &  $\boldsymbol{71.8_{4.6}}^*$ &               $71.4_{5.7}$ &  $\boldsymbol{78.7_{3.6}}^*$ \\
NLU Intent & 68  &                 $69.7_{3.3}$ &  $\boldsymbol{79.7_{1.9}}^*$ &               $72.1_{3.1}$ &  $\boldsymbol{81.9_{1.6}}^*$ &               $74.3_{3.4}$ &  $\boldsymbol{81.6_{2.9}}^*$ \\
BANKING77 & 77  &                 $51.0_{3.4}$ &  $\boldsymbol{63.5_{2.7}}^*$ &               $55.2_{3.3}$ &  $\boldsymbol{69.1_{2.2}}^*$ &               $55.3_{3.5}$ &  $\boldsymbol{70.9_{3.1}}^*$ \\
CLINIC150 & 150 &                 $67.3_{2.7}$ &  $\boldsymbol{75.4_{1.7}}^*$ &                        $68.9_{2.5}$ &                          $\boldsymbol{78.6_{1.8}}^*$ &               $65.7_{5.0}$ &  $\boldsymbol{79.9_{2.1}}^*$ \\
\bottomrule
\end{tabular}}
    \caption{Accuracy results (in \%) for J1-Large, J1-Grande, and J1-Jumbo models with regular ICL in comparison with using \methodshort{} with $B=3$ prompts. The best results for each model and dataset are boldfaced, and `*' is used to indicate that the boldfaced result is statistically better (t-test with p-value $ <0.05$).}
    \label{tab:classification results large and grande }
\end{table*}

\paragraph{Training and test sets}
The performance of \icl~was shown to significantly vary with the choice of training examples~\cite{calibrate}. We followed past work~\cite{calibrate,Fantastically}, randomly sampling 30 sets of training examples from the full training set. We report the mean and standard deviation of performance metrics across these samples. When comparing \methodshort{} method with standard \iclshort, statistically significant differences according to a t-test (p-value $ <0.05$) are marked with \textsuperscript{*}. 
To allow for an extensive set of experiments, we followed prior work and 
randomly subsampled the test sets to contain at most 250 examples \citep{calibrate,Fantastically,pcalibrate}.

\paragraph{Models}
We experiment with 9 LMs of varying sizes: GPT2-Large (0.75B parameters) and GPT2-XL (1.5B) \cite{gpt2}; four LLaMA \cite{llama} models sized 6.7B, 13B, 32.5B and 65.2B;  and three \mbox{Juarassic-1} (J1) models \cite{lieber2021jurassic}: Large (7.5B), Grande (17B), and Jumbo (178B). 
We reduced the number of sampled training sets and the test set  size for the three largest models to 15 and 125, respectively. Notably, J1 and GPT2 models use learned positional embeddings, while LLaMA models use rotary positional embeddings.

\paragraph{Datasets}
Our main focus is classification, and we experiment with 15 different datasets in this category, listed in Appendix~\ref{app:datasets}.
Many of these datasets are used in prior work on in-context learning 
\cite{calibrate,Fantastically,pcalibrate}. 
We additionally experiment with several datasets with a high number of output classes (up to 150), to examine how well our approach works in this setting.
To classify an example in the in-context learning setup, we assign the label using restrictive greedy decoding (see Appendix \ref{app:exp-details}).
We also experiment with another type of tasks, information extraction, and test 4 datasets with a subset of the models (J1-Large and \mbox{J1-Grande}). For these tasks we use greedy decoding at temperature 0 (as in \citet{calibrate}).
For further information about the decoding and formats used for the different types of datasets, see Appendices \ref{app:exp-details} and \ref{app:datasets}.

\begin{table*}[t!]
\small
\centering
\begin{tabular}{lc ll ll ll ll}
\toprule
          &     & \multicolumn{2}{c}{6.7B} & \multicolumn{2}{c}{13B} & \multicolumn{2}{c}{32.5B} & \multicolumn{2}{c}{65B} \\
          \cmidrule(lr){3-4} \cmidrule(lr){5-6} \cmidrule(lr){7-8} \cmidrule(lr){9-10}
         Dataset &  \#  Labels  &                          ICL &               \methodshort{} &                        ICL &               \methodshort{} &                        ICL &               \methodshort{} &                        ICL &               \methodshort{} \\
\midrule
SST-2 & 2   &                 $93.9_{1.2}$ &  $\boldsymbol{94.6_{0.8}}^*$ &  $\boldsymbol{94.5_{0.7}}$ &                 $94.1_{0.7}$ &  $\boldsymbol{94.4_{0.3}}^*$ &                 $93.1_{1.9}$ &  $\boldsymbol{94.1_{0.5}}$ &                 $93.8_{0.5}$ \\
CR & 2   &    $\boldsymbol{93.7_{0.7}}$ &                 $93.6_{0.6}$ &               $92.0_{1.4}$ &    $\boldsymbol{92.2_{0.8}}$ &                 $92.3_{1.0}$ &  $\boldsymbol{93.2_{1.1}}^*$ &               $92.3_{0.9}$ &    $\boldsymbol{92.7_{0.8}}$ \\
RTE & 2   &                 $72.8_{2.6}$ &    $\boldsymbol{73.8_{1.9}}$ &  $\boldsymbol{74.0_{2.6}}$ &                 $73.3_{1.6}$ &    $\boldsymbol{81.0_{1.6}}$ &                 $80.5_{2.2}$ &  $\boldsymbol{78.8_{2.9}}$ &                 $77.5_{2.4}$ \\
Subj & 2   &   $\boldsymbol{72.1_{10.0}}$ &                 $68.9_{9.6}$ &  $\boldsymbol{90.7_{3.9}}$ &                 $89.3_{2.7}$ &                 $90.8_{4.5}$ &    $\boldsymbol{92.1_{3.7}}$ &               $92.1_{4.5}$ &    $\boldsymbol{93.9_{2.6}}$ \\
CB & 3   &    $\boldsymbol{83.2_{4.4}}$ &                 $82.4_{7.1}$ &              $73.4_{10.6}$ &    $\boldsymbol{76.4_{9.1}}$ &    $\boldsymbol{88.5_{2.0}}$ &                 $87.0_{2.2}$ &               $88.5_{1.4}$ &  $\boldsymbol{90.1_{1.6}}^*$ \\
AGNews & 4   &    $\boldsymbol{88.1_{2.3}}$ &                 $87.5_{1.7}$ &               $86.9_{3.0}$ &    $\boldsymbol{87.9_{1.7}}$ &                 $83.4_{3.2}$ &    $\boldsymbol{83.9_{1.2}}$ &               $84.1_{2.0}$ &    $\boldsymbol{84.3_{1.5}}$ \\
YELP & 5   &                $59.0_{10.2}$ &    $\boldsymbol{61.3_{6.0}}$ &               $67.5_{5.4}$ &    $\boldsymbol{69.0_{2.9}}$ &                 $62.3_{5.5}$ &    $\boldsymbol{63.9_{5.8}}$ &               $64.1_{3.3}$ &    $\boldsymbol{65.2_{3.7}}$ \\
SST-5 & 5   &    $\boldsymbol{50.7_{4.7}}$ &                 $48.9_{5.8}$ &               $54.2_{2.3}$ &    $\boldsymbol{55.2_{2.2}}$ &                 $53.2_{3.0}$ &    $\boldsymbol{54.5_{3.0}}$ &               $55.4_{3.2}$ &  $\boldsymbol{58.9_{2.1}}^*$ \\
TREC & 6   &    $\boldsymbol{83.2_{5.6}}$ &                 $82.9_{4.4}$ &               $\boldsymbol{83.7_{3.1}}$ &                 $\boldsymbol{83.7_{2.9}}$ &                 $86.4_{2.6}$ &  $\boldsymbol{89.3_{1.7}}^*$ &               $88.1_{2.2}$ &    $\boldsymbol{89.2_{2.2}}$ \\
DBPedia & 14  &                 $87.9_{6.2}$ &  $\boldsymbol{94.9_{2.8}}^*$ &               $88.3_{6.4}$ &  $\boldsymbol{93.4_{4.2}}^*$ &                 $88.5_{6.7}$ &  $\boldsymbol{94.5_{3.9}}^*$ &               $88.0_{6.2}$ &  $\boldsymbol{96.0_{2.6}}^*$ \\
NLU Scenario & 18  &                 $79.5_{2.8}$ &  $\boldsymbol{82.4_{2.0}}^*$ &               $83.5_{2.2}$ &  $\boldsymbol{87.8_{1.3}}^*$ &                 $84.2_{3.1}$ &  $\boldsymbol{86.9_{1.7}}^*$ &               $86.0_{2.7}$ &  $\boldsymbol{89.1_{1.3}}^*$ \\
TREC Fine & 50  &    $\boldsymbol{54.6_{6.9}}$ &                 $52.6_{5.9}$ &               $55.4_{5.3}$ &  $\boldsymbol{60.1_{5.1}}^*$ &                 $64.0_{6.0}$ &  $\boldsymbol{71.4_{4.1}}^*$ &               $64.4_{4.8}$ &    $\boldsymbol{67.7_{4.1}}$ \\
NLU Intent & 68  &                 $61.6_{4.4}$ &    $\boldsymbol{63.6_{3.6}}$ &               $66.5_{3.9}$ &  $\boldsymbol{72.6_{2.5}}^*$ &                 $70.8_{4.9}$ &  $\boldsymbol{75.7_{3.2}}^*$ &               $73.5_{5.0}$ &  $\boldsymbol{81.1_{3.0}}^*$ \\
BANKING77 & 77  &                 $46.3_{3.8}$ &  $\boldsymbol{53.0_{3.4}}^*$ &               $46.7_{4.0}$ &  $\boldsymbol{56.1_{3.0}}^*$ &                 $51.6_{4.7}$ &  $\boldsymbol{65.4_{3.8}}^*$ &               $54.5_{6.1}$ &  $\boldsymbol{67.8_{2.9}}^*$ \\
CLINIC150 & 150 &  $\boldsymbol{60.9_{2.7}}^*$ &                 $58.2_{3.8}$ &               $63.7_{2.5}$ &  $\boldsymbol{66.0_{2.7}}^*$ &                 $69.4_{4.2}$ &  $\boldsymbol{75.0_{3.9}}^*$ &               $71.8_{2.3}$ &  $\boldsymbol{78.9_{1.8}}^*$ \\
\bottomrule
\end{tabular}        \caption{Accuracy results (in \%) for the four LLaMA models with regular ICL in comparison with using \methodshort{} with $B=3$ prompts. The best results for each model and dataset are boldfaced, and `*' is used to indicate that the boldfaced result is statistically better (t-test with p-value $ <0.05$).}
    \label{tab:llama-models}
\end{table*}

\subsection{Classification Tasks Results}
\label{results-classification}

\paragraph{\methodshort{} enables \icl{} with a large number of classes.}

Tables \ref{tab:classification results large and grande } and \ref{tab:llama-models} show the results for the J1 and LLaMA models, respectively, on various classification tasks, organized by the number of classes. 
With a small number of output classes ($\leq6$), we find small or insignificant differences between \methodshort{} and vanilla  \iclshort{} on J1-Large (7.5B), while with J1-Grande (17B) and J1-Jumbo (178B), \methodshort{} is superior in the majority of cases. However, many of these differences are not statistically significant. A similar pattern is observed with LLaMA models.

Our \methodshort{} method shines in classification tasks with a large number of output classes. With more than 6 classes, \methodshort{} statistically significantly outperforms \iclshort{} in nearly all models and datasets. The average improvement across these datasets is $7.4$, $8.2$, and $8.7$ for J1-Large, J1-Grande, and J1-Jumbo, and $2.3$, $5.3$, $6.7$, and $7.1$ for LLaMA models, in increasing model size. 
Evidently, \emph{the larger the model, the greater the benefit from our method}. This positive scaling behavior of \methodshort{} stands in contrast to prior work attempting to improve ICL \citep{calibrate,Fantastically,pcalibrate}, where improvements to  \mbox{178B-scale} models were smaller than improvements observed in smaller models. 

In Table \ref{tab:gpt2-results} (Appendix \ref{app:gpt2}), we report results with GPT2 models. Although they are smaller than J1 and LLaMA models, we find consistent statistically significant improvements with GPT2-XL (1.5B parameters) in almost all datasets. With GPT2-Large (0.75B), we find improvements in the majority of datasets.

\begin{figure}[t!]
    \includegraphics[clip,width=\columnwidth]{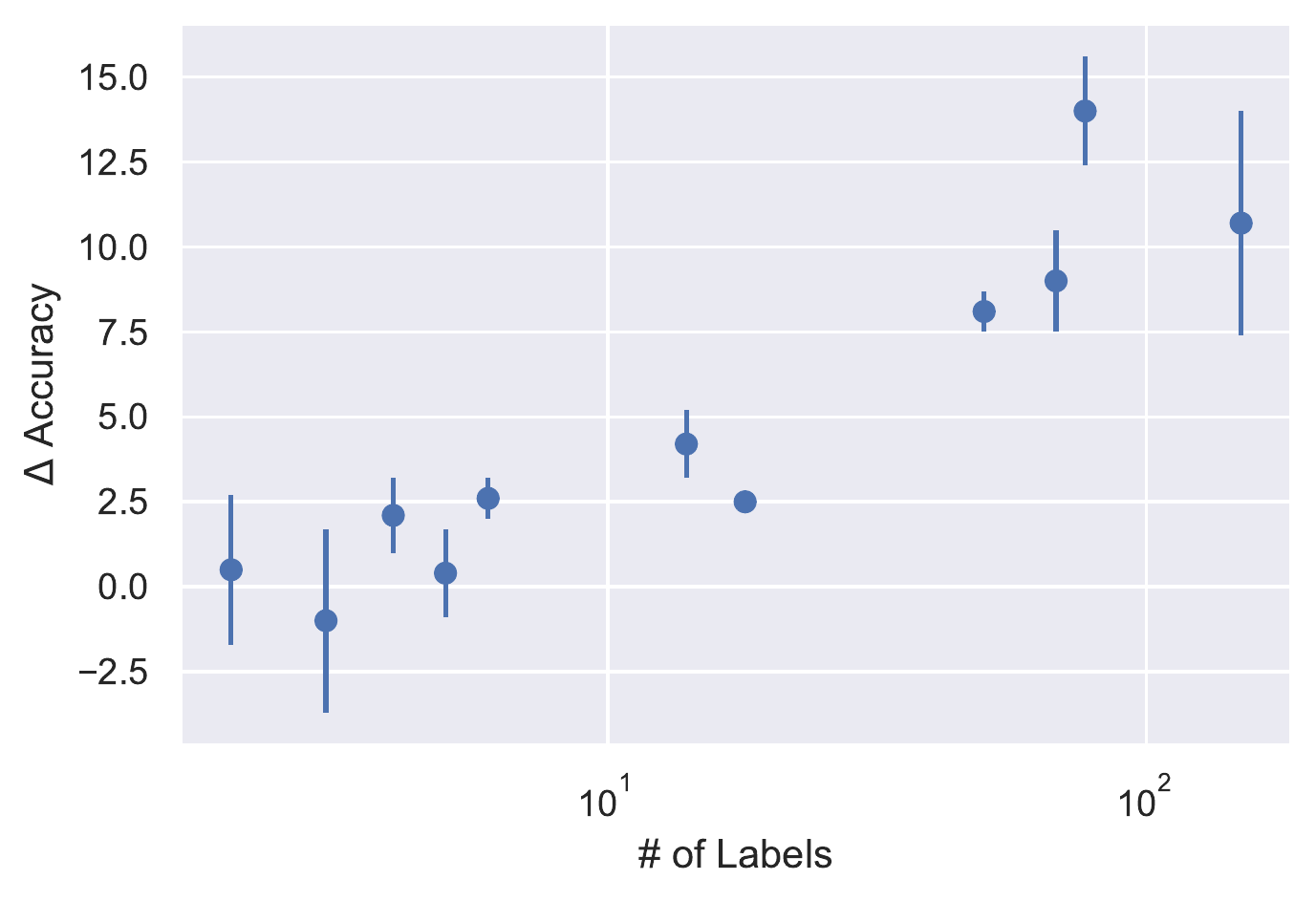}
     \centering
        \caption{Average gains of \methodshort{} vs.\ the \# of labels. Each data point represents the average gain across all datasets and J1 models. 
        There is a a strong positive correlation between the number of unique labels and the gains from \methodshort{}.
        }
        \label{fig:gains-per-n}
\end{figure}

\paragraph{\methodshort{} improves with more classes.}

To examine the relation between the number of output classes and the performance of \methodshort{}, we compute the difference between \methodshort{} and \iclshort{} in each experiment, and average over all datasets (and models) having the same number of classes. 
As Figure \ref{fig:gains-per-n} shows, there is a strong positive correlation between the number of classes and the improvement brought about by \methodshort{} for J1 models (Pearson correlation $r = 0.92$ between the log-number of classes and the average improvement; the slope is 3.02). The correlation for LLaMA models is slightly smaller but still high (Pearson $r = 0.79$).
For datasets with dozens of unique labels---specifically Banking77 \cite{banking77}, NLU Intent \cite{nlu}, and CLINIC150 \cite{clinic150}---we observe improvements of 10--15 points in many cases. Importantly, 
prior \icl{} work has not considered datasets with such a large number of classes, perhaps due to the standard limitation of the context window size.\footnote{An exception is \citet{raft}, who evaluated GPT3 on Banking77 in a limited setting, but obtained poor results.}
We note that in GPT2 models (Table \ref{tab:gpt2-results}, Appendix \ref{app:gpt2}) we do not see a significant correlation between \methodshort{} improvements and the number of classes, but these smaller models tend to struggle with very large numbers of classes.

Comparing results for datasets with different numbers of output classes may be confounded by other factors, such as differences in domain, style, or genre. 
To isolate such effects, we compare results with two datasets, each having both fine-grained and coarse-grained labels: (1) 
The TREC dataset \cite{TREC}, which has 6 coarse-grained and 50 fine-grained classes. (2) NLU \cite{nlu},\footnote{Note that the NLU dataset is also misleadingly known as HWU64; see the 
\href{https://huggingface.co/datasets/nlu_evaluation_data}{Huggingface dataset page} for more details.}  which has 18 scenarios and 68 intents.
From Table \ref{tab:classification results large and grande }, we see that \methodshort{} outperforms standard \iclshort{} by 2.6 and 8.1 points on TREC coarse-grained and fine-grained classification, respectively. Similarly, on NLU coarse- and fine-grained classification, we see average improvements of 2.5 and 9.0 points, respectively.  Similiar trend can be seen in Table \ref{tab:llama-models}.
We conclude that our approach shines especially well when dealing with a large number of output classes.

\paragraph{\methodshort{} makes \icl{} more stable.}
A known limitation of \icl{} is high variance across examples and sensitivity to aspects like the order of examples \cite{Fantastically}. 
Encouragingly, we find that \methodshort{}  reduces such variance: We observe average std values of $3.1$, $2.3$, and $2.6$ for J1-Large, J1-Grande, and J1-Jumbo with \methodshort{}, compared to  $3.9$, $3.4$, and $3.9$ in standard \iclshort{}. Similarly, the average std values for the LLaMA models (ordered by increasing model size) are $4.0$, $2.9$, $2.9$, and $2.2$ for PCW, compared to $4.6$, $3.8$, $3.6$, and $3.2$ for vanilla ICL.

\subsection{Information Extraction Results} \label{sec:icl-multi}

Table \ref{tab:results-ie} shows the results of \iclshort{} and \methodshort{} on information extraction datasets with tasks like airline name extraction or extractive question answering.
These tasks can be considered as classification tasks with an extremely large number of classes, potentially the entire vocabulary or phrases from the vocabulary.
Our approach consistently improves results with both J1-Large and J1-Grande, resulting in statistically significant improvements in almost all cases. 
We also observe smaller standard deviations with \methodshort{} compared to \iclshort{}. 

It is worth noting that prior work has not experimented much with information extraction in an \icl{} setting. \citet{calibrate} reported results with several datasets, but not with extractive question-answering. Our approach seems to allow \icl{} in such cases as well. 

Finally, we tested two multiple-choice QA tasks: OpenBookQA \cite{OpenBookQA2018} and StoryCloze \cite{storycloze}. With our larger model, J1-Grande, \methodshort{} leads to a significant improvement in OpenBookQA and does not significantly improve or worsen over \iclshort{} in other cases.
Details and results of the experiment can be found in Appendix~\ref{app:multiple-choice}.

\begin{table}[t]
    \centering
    \resizebox{\columnwidth}{!}{%
\begin{tabular}{ l ll ll }
\toprule
  & \multicolumn{2}{c}{J1-Large (7.5B)} & \multicolumn{2}{c}{J1-Grande (17B)} \\
  \cmidrule(lr){2-3} \cmidrule(lr){4-5}
   Dataset&           ICL &               \methodshort{} &                        ICL &               \methodshort{} \\
\midrule
ATIS &   $85.6_{5.3}$ &  $\boldsymbol{89.0_{3.0}}^*$ &               $88.0_{4.6}$ &    $\boldsymbol{91.7_{3.1}}^*$ \\
MIT Movies &   $67.9_{2.7}$ &  $\boldsymbol{70.3_{2.5}}^*$ &  ${69.0_{3.9}}$ &                 $\boldsymbol{69.3_{3.3}}$ \\
SQuAD &  $79.2_{2.1}$ &  $\boldsymbol{80.5_{1.4}}^*$ &               $83.8_{2.5}$ &  $\boldsymbol{85.1_{1.4}}^*$ \\
adversarialQA &  $43.0_{2.2}$ &    $\boldsymbol{44.6_{1.5}}^*$ &               $46.4_{2.0}$ &  $\boldsymbol{47.4_{1.8}}$ \\
\bottomrule
\end{tabular}
}
    \caption{PCW improves information extraction (ATIS and MIT Movies are measured with EM, SQuAd and adversarialQA with F1).
    }
    \label{tab:results-ie}
\end{table}

\section{PCW for Question Answering} \label{sec:additional}
In this section, we explore potential usages of PCW in other settings than \icl{}. Specifically, we examined
two question-answering settings where \methodshort{} is expected to help aggregate information from multiple texts. Firstly, we consider the case of question answering based on retrieved documents. Secondly, we experiment with multi-hop reasoning, where the model is required to utilize more than one text while answering a question. 
Importantly, while in Section~\ref{sec:icl} the parallel context windows were used for processing training examples for \iclshort{}, in this section the windows are used for parallel processing of documents related to the test example.

\subsection{Retrieval Based Question Answering} \label{sec:nq}
\paragraph{Setup}

We first experiment with Natural Questions \cite[NQ,][]{kwiatkowski-etal-2019-natural} in an open-book question-answering retrieval setting: Given a question and a set of candidate documents, that may or may not contain the evidence for the question, a model needs to generate a free-text answer.

In the single context window setting (the baseline), we followed the few-shot setup defined by \citet{Lazaridou-Internet-augmented-language-models}: For each question, we retrieved evidence documents from Wikipedia, using a BM25 sparse retriever~\cite{robertson2009probabilistic}. We then prompted the model with in-context training examples of the related task of extracting the answer from a gold evidence document, and concatenated the test question and $N\in\{1,2,4,6,8,10\}$ evidence documents\footnote{Notably, \citet{Lazaridou-Internet-augmented-language-models} used this apparatus only with $N=1$, while we experiment with different values of $N$.}. To fully utilize the context window size, we ``padded'' the prompt with as much in-context training examples as possible.

For \methodshort{}, we followed the setup of a single window while taking advantage of the method's natural ability of parallelization: We increased the number of retrieved documents per question, and divided them between windows. {\emph{E.g.}}, for $N=1$ and 3 parallel context windows ($B=3$), \methodshort{} processes $B\times N=3$ retrieved documents (1 per each window), thus effectively increasing the chance that the correct answer span will be shown to the model in one of the retrieved documents. The metric we used was the standard Exact Match (EM). We refer to Appendix \ref{app:exp-details} for more details.

\begin{figure}[t!]
    \includegraphics[clip,width=\columnwidth]{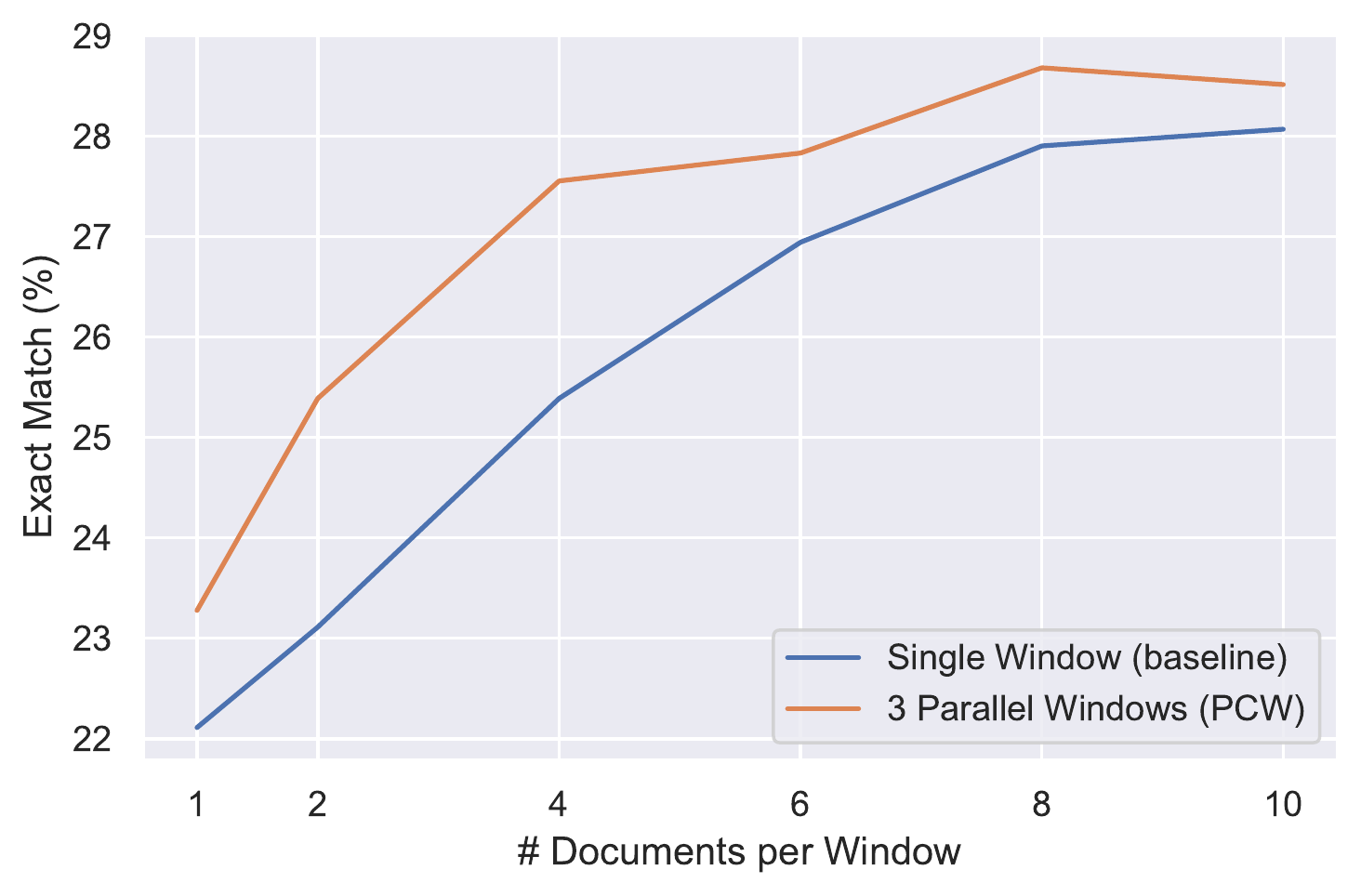}
     \centering
        \caption{
        NQ results for J1 Grande---EM against number of documents in a single window. 
        }
        \label{fig:nq}
\end{figure}

\paragraph{Results}
 Figure~\ref{fig:nq} shows the results for  J1-Grande, when using \methodshort{} compared to the baseline, as a function of the number of candidate documents in a single window.
 In all cases, \methodshort{} performs better than the baseline, demonstrating the benefit of parallel processing of candidate documents. 
 As we increase the number of available retrieved documents, we see an increase in performance for both approaches.
 Similar trend can be seen for J1-Large (see Figure \ref{fig:nq-large} in Appendix).
 Naturally, the performance of this task depends on the probability of retrieving the correct answer. The latter increases in \methodshort{} setting, when the number of processed documents is multiplied by $B=3$.

\begin{table}[t]
    \small
    \centering
    \begin{tabular}{ l rr rr }
\toprule
           & \multicolumn{2}{c}{J1-Large {\small (7.5B)} } & \multicolumn{2}{c}{J1-Grande {\small (17B)}} \\
           \cmidrule(lr){2-3} \cmidrule(lr){4-5}
      Type     &            Seq\  &             PCW\   &                          Seq\ & PCW\    \\
      \midrule

Comparison &  $15.3$ &   $\boldsymbol{21.5}$ &
$20.9$ &  $\boldsymbol{28.7}$ \\
Bridge &   $\boldsymbol{21.6}$ &  $16.5$ &
$\boldsymbol{27.1}$ &  $24.0$ \\
\bottomrule
\end{tabular}
    \caption{
    Zero-shot Exact Match (EM) results on HotpotQA with Sequential and Parallel processing (PCW approach).
    }
    \label{tab:hotpot}
\end{table}

\subsection{Multi-hop Question Answering} \label{sec:hotpot}
\paragraph{Setup}
Finally, we experiment with HotpotQA \cite{yang-etal-2018-hotpotqa}, which requires multi-hop reasoning. Given a question and 10 evidence documents (2 gold and 8 distractors), answering the question requires reasoning over both gold documents. 
HotpotQA includes two question types\footnote{Examples of questions taken from \citet{yang-etal-2018-hotpotqa}}: (a) Questions that refer to a \emph{bridge} entity. For example, to answer the question ``when was the singer of Radiohead born?'', one needs to reason that the singer is ``Thom Yorke'' (the \emph{bridge} entity) and then find his birthday. (b) Questions that rely on a \emph{comparison} between two entities. For example: ``Who has played for more NBA teams, Michael Jordan or Kobe Bryant?''. 
As a baseline, we provide all of the evidences in a random, sequential manner. 
For PCW, we use 5 windows, with 2 evidences in each window.
Since the 10 evidences filled most of the context window of J1 models, we work in a zero-shot setting. The evaluation metric is the standard Exact Match (EM).

\paragraph{Results}
Table \ref{tab:hotpot} shows the results. We break down the results according to the bridge and comparison question types.
Interestingly, \methodshort{}  helps with comparison questions, improving performance over the baseline in both J1-Large and J1-Grande while degrading the performance on bridge questions. 
This disparate behavior can be explained by the kind of processing required to answer the two types of questions. 
In comparison questions, the model can extract the necessary information from the two gold texts independently, making them suitable for \methodshort{}.
For example, to know who played for more NBA teams, the LM needs to extract the number of NBA teams Jordan played for from one text, while extracting the number of NBA teams Bryant played for from another independent text.
In contrast, to answer a bridge question, the processing of each text is conditioned on the other text: When reading a sentence about Thom Yorke's birthplace, we already need to know that Yorke is the Radiohead singer, if we wish to then be able to answer the above question. This makes \methodshort{} less suitable for these types of tasks in its current form, and we leave it as an open direction for how to encode sequential relation between windows (perhaps by some further training).

\section{Related Work}
\subsection{In-Context Learning}

In-context learning has been the subject of extensive research since it was first introduced by \citet{gpt3}. 
For instance, \citet{calibrate} showed that LMs are often miscalibrated. \citet{calibrate} and \citet{pcalibrate} explored ways to overcome this issue by different calibration methods. \citet{Fantastically} observed that few-shot performance  varies significantly depending on the order of examples in the prompt, and proposed a protocol for finding better permutations.
\citet{noisy_channel} proposed a noisy channel approach to boost few-shot performance. 
Our framework is orthogonal and thus complementary to these methods, as we are mainly focused on how to increase the number of examples shown to the model.
Our approach is also more general as it seamlessly supports generative tasks as well. 

\subsection{Expanding the Context Window}
The issue of a limited context window has been the focus of many studies that tried to alleviate the memory footprint of self-attention.
One line of work \citep[][\textit{inter alia}]{bigbird,longT5} suggested using sparse attention to overcome this difficulty. 
\citet{alibi} proposed to encode positional information via relative factors added to attention weights, instead of absolute positional encoding.
Despite the impressive extrapolation abilities of \citet{alibi}, the self-attention cost of such models remains quadratic, making  inference for longer prompts slow and expensive. 
\citet{sled} suggest SLED, an encoder--decoder model for long texts, which encodes short overlapping chunks of the input text, and fuses the information in the decoder, \textit{a-la} Fusion-in-Decoder \cite{izacard2020leveraging}. 
Similarly to our approach, both \citet{izacard2020leveraging} and \citet{sled} employ off-the-shelf architectures, but those methods require further training. Among all mentioned methods, our work is the first that utilizes existing LLMs for longer inputs \textit{without any further training}.

In concurrent work, \citet{hao2022structured} suggest using multiple context windows, while scaling the context tokens' attention weights. We show that large gains can be made without scaling the attention weights, and we demonstrate particularly large gains for tasks with diverse output spaces. 

Moreover, they focus on LLMs with non-learned positional encoding (sinusoidal,~\citealt{attention} and ALIBI,~\citealt{alibi}) and only show results in the ICL setting. In
contrast, we show that \methodshort{} is effective for more common LLMs that have
learned or rotary positional embeddings, and show that \methodshort{} obtains gains both in ICL and in document retrieval settings.

\section{Conclusion and Future Work}

In recent years, a multitude of successful approaches have been proposed for allowing Transformer-based language models to leverage large amounts of text during inference, leading to a variety of dedicated architectures. 
In parallel, however, the mainstream LLM production line of new models with ``regular''---up to several thousand tokens---context window sizes enjoys faster progress in the form of scaling, innovation, and data updating. 

This paper introduced~\textit{\method} (PCW): A simple approach for allowing any off-the-shelf LLM to broaden the scope of text it can access during inference. We showed the effectiveness of~\methodshort~in the framework of~\icl, where access to a context that is larger by a factor of $B$ implies learning from  $B$ times more training examples. 
Our results show that \methodshort~is more effective than the vanilla single context window approach for \icl~over a broad set of multi-class classification tasks, suggesting that \methodshort~could improve \icl~in tasks with diverse input or output spaces.
We also showed promising signals for applying~\methodshort{} for multiple retrieved document reading.

Two key directions of future work strike us as particularly promising. First, by demonstrating that an off-the-shelf LLM can attend to substantially larger quantities of text via~\methodshort, our results motivate further investigation of the \methodshort~method in other settings in which it would be desirable to apply mainstream LLMs over long text sequences.
Second, though our results suggest that \methodshort{}  is effective without further training, we believe that further (short) training of an LLM with parallel context windows could further enhance the abilities demonstrated in this work.

\section*{Limitations}
\label{sec:limitations}
We presented \method{} (\methodshort{}), a simple approach that alleviates context window restrictions for any off-the-shelf LLM, without additional training. We showed the potential of this method on a variety of models and datasets. With that, our method does have some limitations. 

\paragraph{The number of context windows has a limit, and needs to be predetermined.}
Similarly to vanilla in-context learning, the number of examples to include in the prompt must be selected beforehand. For \methodshort{}, it is also required to select the number of context windows, $B$. In this paper, most of the results are for $B=3$. We experiment in Appendix \ref{app:b_values} with the choice of $B$. The results are task dependent, but at a high level we find that there are diminishing returns around $B$ in the range of 5 to 7. We leave further investigation of how to effectively benefit from more windows for future work.

\paragraph{Not effective for all types of tasks.}
As discussed in Section \ref{sec:icl}, \methodshort{} shows impressive gains in \iclshort{} for tasks such as multi-class tasks classification as well as information extraction. However, for some tasks, \methodshort{} does not improve performance. This might indicate that some tasks are not suited for parallel processing. Section \ref{sec:hotpot} demonstrated that \methodshort{} is more suitable for cases where the input text could be divided into few independent inputs, but it remains an open question as to whether tasks, such as long text generation, would benefit from \methodshort{}. 

\section{Acknowledgements}
We thank our colleagues at AI21 Labs for their assistance and advice and the anonymous reviewers for their useful suggestions.

\bibliography{anthology,references}

\begin{thebibliography}{46}
\expandafter\ifx\csname natexlab\endcsname\relax\def\natexlab#1{#1}\fi

\bibitem[{Alex et~al.(2021)Alex, Lifland, Tunstall, Thakur, Maham, Riedel,
  Hine, Ashurst, Sedille, Carlier, Noetel, and Stuhlmüller}]{raft}
Neel Alex, Eli Lifland, Lewis Tunstall, Abhishek Thakur, Pegah Maham, C.~Jess
  Riedel, Emmie Hine, Carolyn Ashurst, Paul Sedille, Alexis Carlier, Michael
  Noetel, and Andreas Stuhlmüller. 2021.
\newblock \href {https://doi.org/10.48550/ARXIV.2109.14076} {Raft: A real-world
  few-shot text classification benchmark}.

\bibitem[{Bartolo et~al.(2020)Bartolo, Roberts, Welbl, Riedel, and
  Stenetorp}]{aqa}
Max Bartolo, Alastair Roberts, Johannes Welbl, Sebastian Riedel, and Pontus
  Stenetorp. 2020.
\newblock \href {https://doi.org/10.1162/tacl\_a\_00338} {Beat the ai:
  Investigating adversarial human annotation for reading comprehension}.
\newblock \emph{Transactions of the Association for Computational Linguistics},
  8:662--678.

\bibitem[{Brown et~al.(2020)Brown, Mann, Ryder, Subbiah, Kaplan, Dhariwal,
  Neelakantan, Shyam, Sastry, Askell, Agarwal, Herbert-Voss, Krueger, Henighan,
  Child, Ramesh, Ziegler, Wu, Winter, Hesse, Chen, Sigler, Litwin, Gray, Chess,
  Clark, Berner, McCandlish, Radford, Sutskever, and Amodei}]{gpt3}
Tom~B. Brown, Benjamin Mann, Nick Ryder, Melanie Subbiah, Jared Kaplan,
  Prafulla Dhariwal, Arvind Neelakantan, Pranav Shyam, Girish Sastry, Amanda
  Askell, Sandhini Agarwal, Ariel Herbert-Voss, Gretchen Krueger, Tom Henighan,
  Rewon Child, Aditya Ramesh, Daniel~M. Ziegler, Jeffrey Wu, Clemens Winter,
  Christopher Hesse, Mark Chen, Eric Sigler, Mateusz Litwin, Scott Gray,
  Benjamin Chess, Jack Clark, Christopher Berner, Sam McCandlish, Alec Radford,
  Ilya Sutskever, and Dario Amodei. 2020.
\newblock \href {https://doi.org/10.48550/ARXIV.2005.14165} {Language models
  are few-shot learners}.

\bibitem[{Casanueva et~al.(2020)Casanueva, Temcinas, Gerz, Henderson, and
  Vulic}]{banking77}
I{\~{n}}igo Casanueva, Tadas Temcinas, Daniela Gerz, Matthew Henderson, and
  Ivan Vulic. 2020.
\newblock \href {https://arxiv.org/abs/2003.04807} {Efficient intent detection
  with dual sentence encoders}.
\newblock In \emph{Proceedings of the 2nd Workshop on NLP for ConvAI - ACL
  2020}.
\newblock Data available at
  https://github.com/PolyAI-LDN/task-specific-datasets.

\bibitem[{Dagan et~al.(2006)Dagan, Glickman, and Magnini}]{RTE}
Ido Dagan, Oren Glickman, and Bernardo Magnini. 2006.
\newblock The pascal recognising textual entailment challenge.
\newblock In \emph{Machine Learning Challenges. Evaluating Predictive
  Uncertainty, Visual Object Classification, and Recognising Tectual
  Entailment}, pages 177--190, Berlin, Heidelberg. Springer Berlin Heidelberg.

\bibitem[{de~Marneffe et~al.(2019)de~Marneffe, Simons, and Tonhauser}]{CB}
Marie-Catherine de~Marneffe, Mandy Simons, and Judith Tonhauser. 2019.
\newblock The commitmentbank: Investigating projection in naturally occurring
  discourse.

\bibitem[{Ding et~al.(2008)Ding, Liu, and Yu}]{CR}
Xiaowen Ding, Bing Liu, and Philip Yu. 2008.
\newblock \href {https://doi.org/10.1145/1341531.1341561} {A holistic
  lexicon-based approach to opinion mining}.
\newblock pages 231--240.

\bibitem[{Guo et~al.(2021)Guo, Ainslie, Uthus, Ontanon, Ni, Sung, and
  Yang}]{longT5}
Mandy Guo, Joshua Ainslie, David Uthus, Santiago Ontanon, Jianmo Ni, Yun-Hsuan
  Sung, and Yinfei Yang. 2021.
\newblock \href {https://doi.org/10.48550/ARXIV.2112.07916} {Longt5: Efficient
  text-to-text transformer for long sequences}.

\bibitem[{Han et~al.(2022)Han, Hao, Dong, Sun, and Wei}]{pcalibrate}
Zhixiong Han, Yaru Hao, Li~Dong, Yutao Sun, and Furu Wei. 2022.
\newblock \href {https://doi.org/10.48550/ARXIV.2205.10183} {Prototypical
  calibration for few-shot learning of language models}.

\bibitem[{Hao et~al.(2022)Hao, Sun, Dong, Han, Gu, and Wei}]{hao2022structured}
Yaru Hao, Yutao Sun, Li~Dong, Zhixiong Han, Yuxian Gu, and Furu Wei. 2022.
\newblock \href {http://arxiv.org/abs/2212.06713} {Structured prompting:
  Scaling in-context learning to 1,000 examples}.

\bibitem[{Hemphill et~al.(1990)Hemphill, Godfrey, and
  Doddington}]{hemphill-etal-1990-atis}
Charles~T. Hemphill, John~J. Godfrey, and George~R. Doddington. 1990.
\newblock \href {https://aclanthology.org/H90-1021} {The {ATIS} spoken language
  systems pilot corpus}.
\newblock In \emph{Speech and Natural Language: Proceedings of a Workshop Held
  at Hidden Valley, {P}ennsylvania, June 24-27,1990}.

\bibitem[{Ivgi et~al.(2022)Ivgi, Shaham, and Berant}]{sled}
Maor Ivgi, Uri Shaham, and Jonathan Berant. 2022.
\newblock \href {https://doi.org/10.48550/ARXIV.2208.00748} {Efficient
  long-text understanding with short-text models}.

\bibitem[{Izacard and Grave(2020)}]{izacard2020leveraging}
Gautier Izacard and Edouard Grave. 2020.
\newblock Leveraging passage retrieval with generative models for open domain
  question answering.
\newblock \emph{arXiv preprint arXiv:2007.01282}.

\bibitem[{Karpukhin et~al.(2020)Karpukhin, O{\u{g}}uz, Min, Lewis, Wu, Edunov,
  Chen, and Yih}]{karpukhin2020dense}
Vladimir Karpukhin, Barlas O{\u{g}}uz, Sewon Min, Patrick Lewis, Ledell Wu,
  Sergey Edunov, Danqi Chen, and Wen-tau Yih. 2020.
\newblock Dense passage retrieval for open-domain question answering.
\newblock \emph{arXiv preprint arXiv:2004.04906}.

\bibitem[{Kwiatkowski et~al.(2019)Kwiatkowski, Palomaki, Redfield, Collins,
  Parikh, Alberti, Epstein, Polosukhin, Devlin, Lee, Toutanova, Jones, Kelcey,
  Chang, Dai, Uszkoreit, Le, and Petrov}]{kwiatkowski-etal-2019-natural}
Tom Kwiatkowski, Jennimaria Palomaki, Olivia Redfield, Michael Collins, Ankur
  Parikh, Chris Alberti, Danielle Epstein, Illia Polosukhin, Jacob Devlin,
  Kenton Lee, Kristina Toutanova, Llion Jones, Matthew Kelcey, Ming-Wei Chang,
  Andrew~M. Dai, Jakob Uszkoreit, Quoc Le, and Slav Petrov. 2019.
\newblock \href {https://doi.org/10.1162/tacl_a_00276} {Natural questions: A
  benchmark for question answering research}.
\newblock \emph{Transactions of the Association for Computational Linguistics},
  7:452--466.

\bibitem[{Larson et~al.(2019)Larson, Mahendran, Peper, Clarke, Lee, Hill,
  Kummerfeld, Leach, Laurenzano, Tang, and Mars}]{clinic150}
Stefan Larson, Anish Mahendran, Joseph~J. Peper, Christopher Clarke, Andrew
  Lee, Parker Hill, Jonathan~K. Kummerfeld, Kevin Leach, Michael~A. Laurenzano,
  Lingjia Tang, and Jason Mars. 2019.
\newblock \href {https://www.aclweb.org/anthology/D19-1131} {An evaluation
  dataset for intent classification and out-of-scope prediction}.
\newblock In \emph{Proceedings of the 2019 Conference on Empirical Methods in
  Natural Language Processing and the 9th International Joint Conference on
  Natural Language Processing (EMNLP-IJCNLP)}.

\bibitem[{Lazaridou et~al.(2022)Lazaridou, Gribovskaya, Stokowiec, and
  Grigorev}]{Lazaridou-Internet-augmented-language-models}
Angeliki Lazaridou, Elena Gribovskaya, Wojciech Stokowiec, and Nikolai
  Grigorev. 2022.
\newblock \href {https://doi.org/10.48550/ARXIV.2203.05115} {Internet-augmented
  language models through few-shot prompting for open-domain question
  answering}.

\bibitem[{Levine et~al.(2022{\natexlab{a}})Levine, Dalmedigos, Ram, Zeldes,
  Jannai, Muhlgay, Osin, Lieber, Lenz, Shalev-Shwartz
  et~al.}]{levine2022standing}
Yoav Levine, Itay Dalmedigos, Ori Ram, Yoel Zeldes, Daniel Jannai, Dor Muhlgay,
  Yoni Osin, Opher Lieber, Barak Lenz, Shai Shalev-Shwartz, et~al.
  2022{\natexlab{a}}.
\newblock Standing on the shoulders of giant frozen language models.
\newblock \emph{arXiv preprint arXiv:2204.10019}.

\bibitem[{Levine et~al.(2022{\natexlab{b}})Levine, Ram, Jannai, Lenz,
  Shalev-Shwartz, Shashua, Leyton-Brown, and Shoham}]{levine2022huge}
Yoav Levine, Ori Ram, Daniel Jannai, Barak Lenz, Shai Shalev-Shwartz, Amnon
  Shashua, Kevin Leyton-Brown, and Yoav Shoham. 2022{\natexlab{b}}.
\newblock \href {https://openreview.net/forum?id=z3Bxu8xNJaF} {Huge frozen
  language models as readers for open-domain question answering}.
\newblock In \emph{ICML 2022 Workshop on Knowledge Retrieval and Language
  Models}.

\bibitem[{Lhoest et~al.(2021)Lhoest, Villanova~del Moral, Jernite, Thakur, von
  Platen, Patil, Chaumond, Drame, Plu, Tunstall, Davison, {\v{S}}a{\v{s}}ko,
  Chhablani, Malik, Brandeis, Le~Scao, Sanh, Xu, Patry, McMillan-Major, Schmid,
  Gugger, Delangue, Matussi{\`e}re, Debut, Bekman, Cistac, Goehringer, Mustar,
  Lagunas, Rush, and Wolf}]{datasets}
Quentin Lhoest, Albert Villanova~del Moral, Yacine Jernite, Abhishek Thakur,
  Patrick von Platen, Suraj Patil, Julien Chaumond, Mariama Drame, Julien Plu,
  Lewis Tunstall, Joe Davison, Mario {\v{S}}a{\v{s}}ko, Gunjan Chhablani,
  Bhavitvya Malik, Simon Brandeis, Teven Le~Scao, Victor Sanh, Canwen Xu,
  Nicolas Patry, Angelina McMillan-Major, Philipp Schmid, Sylvain Gugger,
  Cl{\'e}ment Delangue, Th{\'e}o Matussi{\`e}re, Lysandre Debut, Stas Bekman,
  Pierric Cistac, Thibault Goehringer, Victor Mustar, Fran{\c{c}}ois Lagunas,
  Alexander Rush, and Thomas Wolf. 2021.
\newblock \href {http://arxiv.org/abs/2109.02846} {Datasets: A community
  library for natural language processing}.
\newblock In \emph{Proceedings of the 2021 Conference on Empirical Methods in
  Natural Language Processing: System Demonstrations}, pages 175--184, Online
  and Punta Cana, Dominican Republic. Association for Computational
  Linguistics.

\bibitem[{Li and Roth(2002)}]{TREC}
Xin Li and Dan Roth. 2002.
\newblock \href {https://www.aclweb.org/anthology/C02-1150} {Learning question
  classifiers}.
\newblock In \emph{{COLING} 2002: The 19th International Conference on
  Computational Linguistics}.

\bibitem[{Lieber et~al.(2021)Lieber, Sharir, Lenz, and
  Shoham}]{lieber2021jurassic}
Opher Lieber, Or~Sharir, Barak Lenz, and Yoav Shoham. 2021.
\newblock Jurassic-1: Technical details and evaluation.
\newblock \emph{White Paper. AI21 Labs}.

\bibitem[{Lin et~al.(2021)Lin, Ma, Lin, Yang, Pradeep, and Nogueira}]{Pyserini}
Jimmy Lin, Xueguang Ma, Sheng-Chieh Lin, Jheng-Hong Yang, Ronak Pradeep, and
  Rodrigo Nogueira. 2021.
\newblock {Pyserini}: A {Python} toolkit for reproducible information retrieval
  research with sparse and dense representations.
\newblock In \emph{Proceedings of the 44th Annual International ACM SIGIR
  Conference on Research and Development in Information Retrieval (SIGIR
  2021)}, pages 2356--2362.

\bibitem[{Liu et~al.(2012)Liu, Cyphers, Pasupat, McGraw, and
  Glass}]{Liu2012ACM}
Jingjing Liu, D.~Scott Cyphers, Panupong Pasupat, Ian McGraw, and James~R.
  Glass. 2012.
\newblock A conversational movie search system based on conditional random
  fields.
\newblock In \emph{Interspeech}.

\bibitem[{Lu et~al.(2021)Lu, Bartolo, Moore, Riedel, and
  Stenetorp}]{Fantastically}
Yao Lu, Max Bartolo, Alastair Moore, Sebastian Riedel, and Pontus Stenetorp.
  2021.
\newblock \href {https://doi.org/10.48550/ARXIV.2104.08786} {Fantastically
  ordered prompts and where to find them: Overcoming few-shot prompt order
  sensitivity}.

\bibitem[{Mihaylov et~al.(2018)Mihaylov, Clark, Khot, and
  Sabharwal}]{OpenBookQA2018}
Todor Mihaylov, Peter Clark, Tushar Khot, and Ashish Sabharwal. 2018.
\newblock Can a suit of armor conduct electricity? a new dataset for open book
  question answering.
\newblock In \emph{EMNLP}.

\bibitem[{Min et~al.(2021)Min, Lewis, Hajishirzi, and
  Zettlemoyer}]{noisy_channel}
Sewon Min, Mike Lewis, Hannaneh Hajishirzi, and Luke Zettlemoyer. 2021.
\newblock \href {https://doi.org/10.48550/ARXIV.2108.04106} {Noisy channel
  language model prompting for few-shot text classification}.

\bibitem[{Mostafazadeh et~al.(2017)Mostafazadeh, Roth, Louis, Chambers, and
  Allen}]{storycloze}
Nasrin Mostafazadeh, Michael Roth, Annie Louis, Nathanael Chambers, and James
  Allen. 2017.
\newblock Lsdsem 2017 shared task: The story cloze test.
\newblock In \emph{Proceedings of the 2nd Workshop on Linking Models of
  Lexical, Sentential and Discourse-level Semantics}, pages 46--51.

\bibitem[{Pang and Lee(2004)}]{subj}
Bo~Pang and Lillian Lee. 2004.
\newblock \href {http://arxiv.org/abs/cs.CL/0409058} {A sentimental education:
  Sentiment analysis using subjectivity summarization based on minimum cuts}.
\newblock \emph{CoRR}, cs.CL/0409058.

\bibitem[{Press et~al.(2022)Press, Smith, and Lewis}]{alibi}
Ofir Press, Noah Smith, and Mike Lewis. 2022.
\newblock \href {https://openreview.net/forum?id=R8sQPpGCv0} {Train short, test
  long: Attention with linear biases enables input length extrapolation}.
\newblock In \emph{International Conference on Learning Representations}.

\bibitem[{Press et~al.(2021)Press, Smith, and Lewis}]{press2021train}
Ofir Press, Noah~A Smith, and Mike Lewis. 2021.
\newblock Train short, test long: Attention with linear biases enables input
  length extrapolation.
\newblock \emph{arXiv preprint arXiv:2108.12409}.

\bibitem[{Radford et~al.(2019)Radford, Wu, Child, Luan, Amodei, and
  Sutskever}]{gpt2}
Alec Radford, Jeff Wu, Rewon Child, David Luan, Dario Amodei, and Ilya
  Sutskever. 2019.
\newblock Language models are unsupervised multitask learners.

\bibitem[{{Rajpurkar} et~al.(2016){Rajpurkar}, {Zhang}, {Lopyrev}, and
  {Liang}}]{squad}
Pranav {Rajpurkar}, Jian {Zhang}, Konstantin {Lopyrev}, and Percy {Liang}.
  2016.
\newblock \href {http://arxiv.org/abs/1606.05250} {{SQuAD: 100,000+ Questions
  for Machine Comprehension of Text}}.
\newblock \emph{arXiv e-prints}, page arXiv:1606.05250.

\bibitem[{Robertson et~al.(2009)Robertson, Zaragoza
  et~al.}]{robertson2009probabilistic}
Stephen Robertson, Hugo Zaragoza, et~al. 2009.
\newblock The probabilistic relevance framework: Bm25 and beyond.
\newblock \emph{Foundations and Trends{\textregistered} in Information
  Retrieval}, 3(4):333--389.

\bibitem[{Shaham et~al.(2022)Shaham, Segal, Ivgi, Efrat, Yoran, Haviv, Gupta,
  Xiong, Geva, Berant et~al.}]{shaham2022scrolls}
Uri Shaham, Elad Segal, Maor Ivgi, Avia Efrat, Ori Yoran, Adi Haviv, Ankit
  Gupta, Wenhan Xiong, Mor Geva, Jonathan Berant, et~al. 2022.
\newblock Scrolls: Standardized comparison over long language sequences.
\newblock \emph{arXiv preprint arXiv:2201.03533}.

\bibitem[{Socher et~al.(2013)Socher, Perelygin, Wu, Chuang, Manning, Ng, and
  Potts}]{sst}
Richard Socher, Alex Perelygin, Jean Wu, Jason Chuang, Christopher~D. Manning,
  Andrew Ng, and Christopher Potts. 2013.
\newblock \href {https://www.aclweb.org/anthology/D13-1170} {Recursive deep
  models for semantic compositionality over a sentiment treebank}.
\newblock In \emph{Proceedings of the 2013 Conference on Empirical Methods in
  Natural Language Processing}, pages 1631--1642, Seattle, Washington, USA.
  Association for Computational Linguistics.

\bibitem[{Su et~al.(2022)Su, Lu, Pan, Murtadha, Wen, and Liu}]{roformer}
Jianlin Su, Yu~Lu, Shengfeng Pan, Ahmed Murtadha, Bo~Wen, and Yunfeng Liu.
  2022.
\newblock \href {http://arxiv.org/abs/2104.09864} {Roformer: Enhanced
  transformer with rotary position embedding}.

\bibitem[{Tay et~al.(2020)Tay, Dehghani, Abnar, Shen, Bahri, Pham, Rao, Yang,
  Ruder, and Metzler}]{tay2020long}
Yi~Tay, Mostafa Dehghani, Samira Abnar, Yikang Shen, Dara Bahri, Philip Pham,
  Jinfeng Rao, Liu Yang, Sebastian Ruder, and Donald Metzler. 2020.
\newblock Long range arena: A benchmark for efficient transformers.
\newblock \emph{arXiv preprint arXiv:2011.04006}.

\bibitem[{Touvron et~al.(2023)Touvron, Lavril, Izacard, Martinet, Lachaux,
  Lacroix, Rozière, Goyal, Hambro, Azhar, Rodriguez, Joulin, Grave, and
  Lample}]{llama}
Hugo Touvron, Thibaut Lavril, Gautier Izacard, Xavier Martinet, Marie-Anne
  Lachaux, Timothée Lacroix, Baptiste Rozière, Naman Goyal, Eric Hambro,
  Faisal Azhar, Aurelien Rodriguez, Armand Joulin, Edouard Grave, and Guillaume
  Lample. 2023.
\newblock \href {http://arxiv.org/abs/2302.13971} {Llama: Open and efficient
  foundation language models}.

\bibitem[{Vaswani et~al.(2017)Vaswani, Shazeer, Parmar, Uszkoreit, Jones,
  Gomez, Kaiser, and Polosukhin}]{attention}
Ashish Vaswani, Noam Shazeer, Niki Parmar, Jakob Uszkoreit, Llion Jones,
  Aidan~N. Gomez, Lukasz Kaiser, and Illia Polosukhin. 2017.
\newblock \href {http://arxiv.org/abs/1706.03762} {Attention is all you need}.
\newblock \emph{CoRR}, abs/1706.03762.

\bibitem[{Xingkun~Liu and Rieser(2019)}]{nlu}
Pawel~Swietojanski Xingkun~Liu, Arash~Eshghi and Verena Rieser. 2019.
\newblock \href {http://www.xx.xx/xx/} {Benchmarking natural language
  understanding services for building conversational agents}.
\newblock In \emph{Proceedings of the Tenth International Workshop on Spoken
  Dialogue Systems Technology (IWSDS)}, pages xxx--xxx, Ortigia, Siracusa (SR),
  Italy. Springer.

\bibitem[{Yang et~al.(2018)Yang, Qi, Zhang, Bengio, Cohen, Salakhutdinov, and
  Manning}]{yang-etal-2018-hotpotqa}
Zhilin Yang, Peng Qi, Saizheng Zhang, Yoshua Bengio, William Cohen, Ruslan
  Salakhutdinov, and Christopher~D. Manning. 2018.
\newblock \href {https://doi.org/10.18653/v1/D18-1259} {{H}otpot{QA}: A dataset
  for diverse, explainable multi-hop question answering}.
\newblock In \emph{Proceedings of the 2018 Conference on Empirical Methods in
  Natural Language Processing}, pages 2369--2380, Brussels, Belgium.
  Association for Computational Linguistics.

\bibitem[{Zaheer et~al.(2020)Zaheer, Guruganesh, Dubey, Ainslie, Alberti,
  Ontanon, Pham, Ravula, Wang, Yang, and Ahmed}]{bigbird}
Manzil Zaheer, Guru Guruganesh, Avinava Dubey, Joshua Ainslie, Chris Alberti,
  Santiago Ontanon, Philip Pham, Anirudh Ravula, Qifan Wang, Li~Yang, and Amr
  Ahmed. 2020.
\newblock \href {https://doi.org/10.48550/ARXIV.2007.14062} {Big bird:
  Transformers for longer sequences}.

\bibitem[{Zhang et~al.(2015{\natexlab{a}})Zhang, Zhao, and
  LeCun}]{many-classifications}
Xiang Zhang, Junbo Zhao, and Yann LeCun. 2015{\natexlab{a}}.
\newblock \href
  {https://proceedings.neurips.cc/paper/2015/file/250cf8b51c773f3f8dc8b4be867a9a02-Paper.pdf}
  {Character-level convolutional networks for text classification}.
\newblock In \emph{Advances in Neural Information Processing Systems},
  volume~28. Curran Associates, Inc.

\bibitem[{Zhang et~al.(2015{\natexlab{b}})Zhang, Zhao, and LeCun}]{agnews}
Xiang Zhang, Junbo~Jake Zhao, and Yann LeCun. 2015{\natexlab{b}}.
\newblock Character-level convolutional networks for text classification.
\newblock In \emph{NIPS}.

\bibitem[{Zhao et~al.(2021)Zhao, Wallace, Feng, Klein, and Singh}]{calibrate}
Tony~Z. Zhao, Eric Wallace, Shi Feng, Dan Klein, and Sameer Singh. 2021.
\newblock \href {https://doi.org/10.48550/ARXIV.2102.09690} {Calibrate before
  use: Improving few-shot performance of language models}.

\end{thebibliography}
\bibliographystyle{acl_natbib}

\appendix 
\section{Experimental Details} \label{app:exp-details}
\subsection{\methodshort{} Implementation Details}
\paragraph{Handling context windows of various lengths} Section \ref{sec:method} thoroughly describes \methodshort{} method for cases where each window has the same number of tokens. Throughout all our experiments, this was rarely the case. We considered two variations of \methodshort{} to handle these cases. The first was whether to use left or right indentation of the windows, meaning whether all of the windows should begin or end in the same position id. To avoid any discontinuity in the assignment of position ids, it is also possible to pad the windows with some dummy tokens (\eg, new line). Left indentation was found to be the most preferred option in \iclshort{} setting, while padding did not appear to be significant. For that reason, and considering the simplicity of this solution, we chose to use left indentation in all of our experiments. It is important to note that in the \methodshort{} implementation, all the windows and the task tokens attend to a single shared BOS token. We found that having multiple BOS tokens negatively affected our results.

\paragraph{Splitting the inputs into windows}
For the experiments described in Section \ref{sec:icl}, we assigned an equal number of $n_{max}$ samples per window, and only attempted to balance the lengths of the windows by greedily switching long and short samples between windows.  $n_{max}$ was calculated according to the following formula:
\begin{equation}
    n_{max} = 	\lfloor \dfrac{N-T_{max}}{D_{90}} \rfloor
\end{equation}
where $N$ is the context window size, $T_{max}$ is the length of longest test sample and $D_{90}$ is the $90$th percentile of the train samples' lengths. To avoid unwanted effects due to outliers, we removed the longest percentile of train and test samples.

In the experiments described in Section \ref{sec:nq}, we divided the documents according to the retriever's ranking, so that the last document in each window would have the highest ranking in the window. It should be noted that the training examples were not parallelized. The same randomly chosen examples were used for both baseline and \methodshort{}, and new examples were drawn for each test sample. For the experiment described in Section \ref{sec:hotpot}, the division between windows was random.

\subsection{Evaluation Details}
\paragraph{Classification}\label{par: classification_eval}
A common way to evaluate models in the in-context learning setup is to iterate over all possible labels for each test sample and check which label receives the highest probability according to the LM. This approach is problematic where a large number of classes is present, especially when some class names are split into multiple tokens. To save computational costs, we implemented constrained greedy decoding, at each step allowing only tokens that could result in a valid label. It is important to acknowledge that this evaluation method could result in slightly different performance for both the \iclshort{} baseline and for the \methodshort{}  approach. However, since most of the labels only contained few tokens, and the first token is usually quite indicative to the nature of the label, this effect should be minor.

\paragraph{Information extraction}
The LMs' predictions for the information extraction tasks were generated with greedy decoding at temperature 0, similar to \citet{calibrate}. We used Exact Match (EM) or F1 as the metric of choice for the extraction tasks.

\paragraph{Computational cost}
As discussed in the beginning of this appendix, we used restrictive decoding for the majority of the experiments in the paper. This usage greatly reduced the computational cost of our experiments: Most classification tasks were preformed in 1-4 GPU hours for all models (besides experiments with J1-Jumbo, which lasted roughly 10-50 GPU hours per experiment).
The experiments described in Section \ref{sec:icl-multi} and Section \ref{sec:additional} took up to 20 GPU hours each.

\section{Datasets Information} \label{app:datasets}
\subsection{Overview}
We used 15 different datasets for our classification experiments:
SST-2 \cite{sst}, CR \cite{CR}, RTE \cite{RTE}, Subj \cite{subj}, CB \cite{CB}, AGNews \cite{agnews}, SST-5 \cite{sst}, YELP \cite{many-classifications},TREC \cite{TREC}, DBPedia \cite{many-classifications}, NLU \cite{nlu}, BANKING77 \cite{banking77} and CLINIC150 \cite{clinic150}. TREC and NLU datasets were used with both fine and coarse grained labels.
The different formats used in all of tasks, as well as the values of $n_{max}$ for the different models, can be found in Table \ref{tab:prompts}. We have also used 6 more datasets from extraction and multiple-choice domains, which were only evaluated with J1 models: 

\begin{itemize}
  \item ATIS airlines \cite{calibrate}; $n_{max}=67$.
  \item MIT Movie Genre \cite{calibrate}; $n_{max}=54$. 
  \item SQuAD \cite{squad}; $n_{max}=8$.
  \item adversarialQA\cite{aqa}; $n_{max}=8$. 
  \item OpenBookQA \cite{OpenBookQA2018}; $n_{max}=87$.
  \item StoryCloze\cite{storycloze}; $n_{max} = 44$.
\end{itemize}

For Section \ref{sec:additional} we used Natural Questions \cite{kwiatkowski-etal-2019-natural} and HotpotQA \cite{yang-etal-2018-hotpotqa} datasets.
All datasets were evaluated with the standard test set or validation set in the absence of a public test set. As described in Section \ref{sec:icl}, we subsampled all test sets for the \iclshort{} experiments. In Natural Questions dataset, we used half of the test set (its original size was 3610 samples) to speed up evaluation. We used the full HotpotQA validation set, containing 7405 samples.
The datasets are all in English.
 
The majority of the datasets can be found in the Huggingface Datasets package \cite{datasets}, apart from the information extraction tasks ATIS airlines \cite{hemphill-etal-1990-atis} and MIT movie genre \cite{Liu2012ACM}, which were taken from \citet{calibrate}, and Natural Questions \cite{kwiatkowski-etal-2019-natural} which was loaded and incorporated with retrieved documents using Pyserini \cite{Pyserini}. Since loading the training set via Pyserini is not currently a built-in option, we used the validation set of Natural Questions as an effective train set. We found this decision reasonable since we only used the training set for few-shot prompting, and we did not optimize any parameters using the validation set.

We have tried our best to track the licenses of the datasets used in this work. The license information that we have managed to find is as follows: SST-2, RTE, SST-5, NLU Scenario, NLU Intent, BANKING77 and SQuAD---CC-BY 4.0, adversarialQA---CC-BY-SA 4.0, DBPedia---CC-BY-SA 3.0.

\subsection{Preprocessing and Formatting}
In all ICL experiments, we used only pairs of inputs and expected outputs, without any instructions. For the classification datasets, we mainly used formats found in \citet{Fantastically} when applicable. For  extraction and multi-choice datasets, we used the formats from \citet{gpt3}. We generated new formats for classification datasets with dozens of labels, which are rarely used in few-shot setting. The formats were based on wordings and labels used in HuggingFace, with minor modifications to make the formats more similar to natural language (\eg, replacing `\_' with spaces in label names). 
Details of the classification prompts can be found in Table \ref{tab:prompts}.
Experiments from Section \ref{sec:additional} were formatted similarly to the work done by \cite{Lazaridou-Internet-augmented-language-models}. Their prompts formats are presented in Table \ref{tab:nq-hotpotqa-format}.

\section{The Effect of the Number of Context Windows on Performance}
\label{app:b_values}
When using \methodshort{} for \iclshort{}, the number of parallel context windows ($B$) affects the number of in-context training examples. We used $B={2,3,4}$ in preliminary experiments, and saw that for classification tasks, the optimal choice of $B$ depends on the number of unique labels in the task. We observed that the performance for tasks with a high number of classes was improved when we increased $B$, while the optimal choice of $B$ for tasks with few classes tended to be 1 or 2 (See Tables \ref{tab:large-diff-ks},\ref{tab:grande-diff-ks} and \ref{tab:Jumbo-diff-ks}). For simplicity, We chose to display results for $B=3$ in all of the main experiments.

Nevertheless, we were curious to see how far we could push the number of parallel context windows before the model stopped benefiting from them. We used a representative subset of three datasets with a varying number of classes, and increased $B$ from 1 to 8. The number of training sets and the size of test set for those experiments were set on 15 and 125 respectively.

\begin{figure*}[t]
    \includegraphics[clip,width=\textwidth]{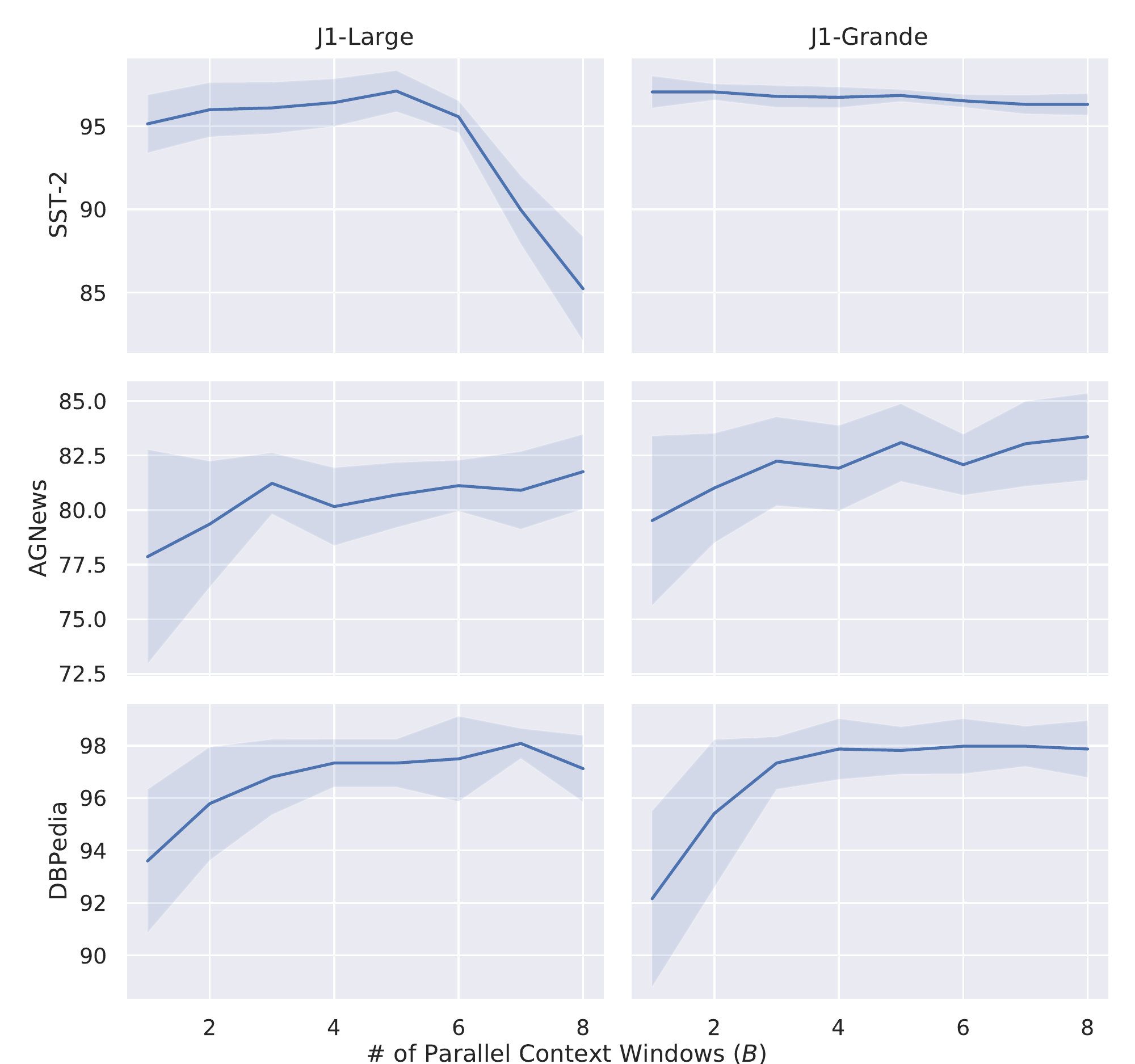}
     \centering
        \caption{Multi-class tasks benefit from increased context windows, but simpler tasks with fewer classes may suffer from a decrease in performance.}
        \label{fig:n-pcw}
\end{figure*}

As seen in Figure \ref{fig:n-pcw}, when the number of context windows is increased, datasets with a large number of classes, such as AGNews and DBPedia (with 4 and 14 labels, respectively), continue to improve (with a convergence at around $B=6$). Hence, PCW can achieve even greater improvements by optimizing $B$ per dataset. Increasing the number of context windows, however, seemed to harm the performance of SST-2.

Identifying which tasks benefit from large parallel data processing would be an interesting research direction in the future. For now, we recommend choosing an optimal $B$ on the development set (if available) for best results. In the absence of a development set, a conservative choice, such as $B=3$, may be beneficial. It is possible to investigate the behaviour of \methodshort{} with a larger number of windows, but we find it irrelevant for most practical cases of \iclshort{}, where an extremely large number of training samples would allow finetuning a model. We leave exploring this issue for future work.

\section{Additional Results}\label{app:add_results}
\subsection{Replication Experiments with GPT2}
\label{app:gpt2}

Table \ref{tab:gpt2-results} presents replication of the results shown in Table \ref{tab:classification results large and grande } for GPT2-Large and GPT2-XL models.
A qualitative inspection of errors in these experiments suggested that vanilla \iclshort{} fails more in examples where the test label appears earlier in the prompt.
Since \methodshort{} allows more context windows, it more often shows a training example with the test label towards the end of one of the windows. 
We evaluated GPT2-Large performance on the AGNews dataset and discovered that
\methodshort{} shows a training example with the test label in a closer location to the test example 62\% of the time. In those cases, \methodshort{} outperforms \iclshort{} by 19.4\%, compared to a margin of roughly 10\% for the entire test set. This analysis suggests that \methodshort{} provides a solution to the recency bias noted by \citet{calibrate}.

\begin{figure}[t!]
    \includegraphics[clip,width=\columnwidth]{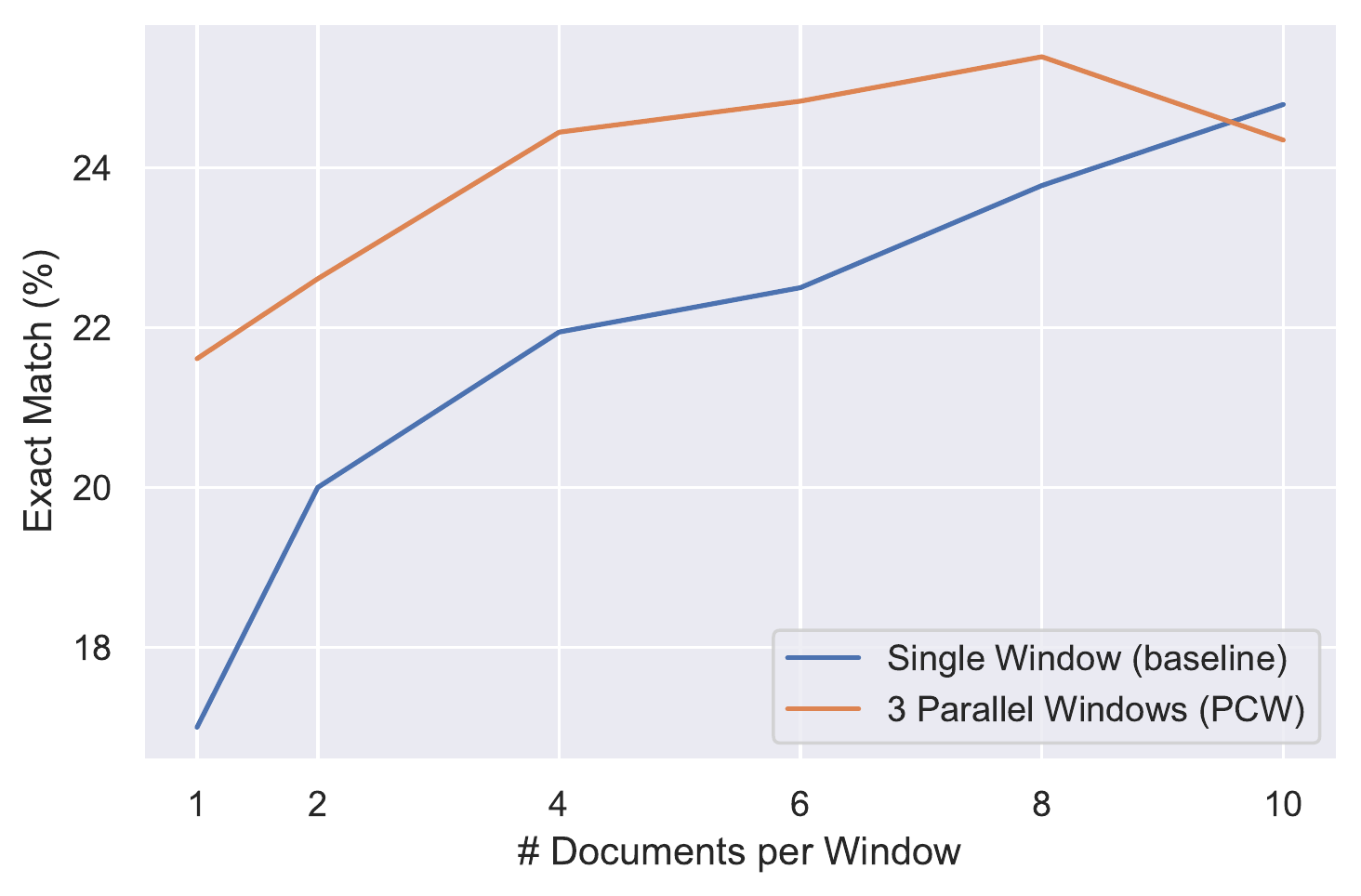}
     \centering
        \caption{NQ results for J1 Large---EM against number of documents in a single window.}
        \label{fig:nq-large}
\end{figure}

\subsection{Multiple-Choice QA}
\label{app:multiple-choice}
In addition to our in-depth investigation of \methodshort{}, we have experimented with two multiple-choice QA datasets OpenBookQA \cite{OpenBookQA2018} and StoryCloze \cite{storycloze} under ICL setting.
We formatted and evaluated the tasks as in \citet{gpt3}, by providing few-shot examples with the correct completion followed by an example of context only, and comparing the average per-token LM likelihood of each possible completion. We did not use the calibration from \citet{gpt3}. We used the same setup as described in Section \ref{sec:setup}, with the exception of reducing the number of sampled training sets and the test set size used for J1-Grande in the OpenBookQA experiment to 15 and 125, respectively.

The results shown in Table \ref{tab:results-multiple} show that increasing the number of examples of in-context training under the \methodshort{} setting improved the performance of J1-Grande in the OpenBookQA task, but did not significantly affect the other scenarios.
Based on this observation, it seems that \methodshort{} has the potential of providing gains for multiple-choice tasks in specific scenarios, but further analysis should be made based on more datasets to better understand it. We leave this for future work.

\begin{table}[t]
\small
\resizebox{\columnwidth}{!}{%
    \centering
    \begin{tabular}{ lc ll ll }
\toprule
           & \multicolumn{2}{c}{J1-Large (7.5B)} & \multicolumn{2}{c}{J1-Grande (17B)} \\
           \cmidrule(lr){2-3} \cmidrule(lr){4-5}
      Dataset     &            ICL &             \methodshort{}   &                          ICL & \methodshort{}   \\
\midrule
OpenBookQA &   $46.0_{1.5}$ &  $\boldsymbol{46.6_{1.0}}$ &  $51.6_{2.2}$ &   $\boldsymbol{54.2_{1.7}^*}$ \\
StoryCloze &  $84.2_{1.0}$ &               $\boldsymbol{84.3_{0.8}}$ &  $\boldsymbol{84.7_{0.9}}$ &   $84.6_{1.0}$ \\
\bottomrule
\end{tabular}}
    \caption{Results for J1-Large and J1-Grande models for  Multiple Choice datasets.}
    \label{tab:results-multiple}
\end{table}

\begin{table*}[t!]
\small
    \centering
\begin{tabular}{  lc  ll ll  }
\toprule
          &     & \multicolumn{2}{c}{GPT2-Large(0.75B)} & \multicolumn{2}{c}{GPT2-XL(1.5B)} \\
          \cmidrule(lr){3-4} \cmidrule(lr){5-6}
         Dataset &\# Labels &                         ICL &                       \methodshort{} &                        ICL &                       \methodshort{} \\
\midrule
SST-2 & 2   &                $80.5_{11.4}$ &  $\boldsymbol{85.5_{5.7}}^*$ &               $90.7_{3.8}$ &  $\boldsymbol{93.0_{2.1}}^*$ \\
CR & 2   &                 $81.3_{6.3}$ &    $\boldsymbol{83.8_{4.7}}$ &               $79.4_{6.0}$ &  $\boldsymbol{82.9_{3.4}}^*$ \\
RTE & 2   &                 $53.0_{2.5}$ &    $\boldsymbol{53.5_{1.9}}$ &               $55.4_{2.7}$ &    $\boldsymbol{55.5_{2.0}}$ \\
Subj & 2   &                $67.4_{12.3}$ &   $\boldsymbol{69.5_{11.8}}$ &              $68.0_{10.8}$ &    $\boldsymbol{68.6_{6.7}}$ \\
CB & 3   &    $\boldsymbol{45.3_{4.7}}$ &                 $44.4_{4.2}$ &  $\boldsymbol{53.5_{9.2}}$ &                 $51.9_{7.0}$ \\
AGNews & 4   &                $61.9_{12.9}$ &  $\boldsymbol{72.5_{7.0}}^*$ &              $68.0_{12.4}$ &  $\boldsymbol{80.0_{3.5}}^*$ \\
SST-5 & 5   &                 $41.1_{4.6}$ &  $\boldsymbol{44.7_{4.4}}^*$ &               $37.1_{7.9}$ &  $\boldsymbol{43.3_{5.9}}^*$ \\
TREC & 6   &                 $55.6_{8.3}$ &    $\boldsymbol{57.7_{4.9}}$ &               $48.4_{4.7}$ &    $\boldsymbol{48.6_{2.6}}$ \\
DBPedia & 14  &                $63.1_{18.9}$ &  $\boldsymbol{80.7_{5.3}}^*$ &              $77.2_{10.5}$ &  $\boldsymbol{87.3_{3.9}}^*$ \\
NLU Scenario & 18  &  $\boldsymbol{37.0_{6.1}}^*$ &                 $31.4_{3.7}$ &               $47.5_{8.0}$ &  $\boldsymbol{52.9_{6.1}}^*$ \\
TREC Fine & 50  &                 $30.3_{7.8}$ &  $\boldsymbol{33.6_{4.0}}^*$ &               $36.8_{6.4}$ &  $\boldsymbol{39.5_{2.8}}^*$ \\
NLU Intent & 68  &                 $24.3_{5.4}$ &  $\boldsymbol{28.1_{4.4}}^*$ &               $30.2_{5.2}$ &  $\boldsymbol{38.9_{4.5}}^*$ \\
BANKING77 & 77  &    $\boldsymbol{29.3_{5.3}}$ &                 $28.5_{4.0}$ &               $30.9_{4.0}$ &  $\boldsymbol{33.7_{3.2}}^*$ \\
CLINIC150 & 150 &                 $44.2_{3.2}$ &    $\boldsymbol{45.4_{1.8}}$ &               $46.9_{2.5}$ &  $\boldsymbol{48.7_{1.9}}^*$ \\
\bottomrule
\end{tabular}
       \caption{Accuracy results for GPT2-Large and GPT2-XL models with regular ICL in comparison with using \methodshort{} with $B=3$ prompts. The results mirror the results found in Table \ref{tab:classification results large and grande } and use the same format, with the exception of YELP dataset that had $n_{max} = 0$ for GPT2 models.}
    \label{tab:gpt2-results}
\end{table*}

\begin{table*}[t!]
\small
\centering
\begin{tabular}{lc llll}
\toprule
          Dataset &  \# Labels   &           ICL &                 \methodshort{} ($B=2$) &                 \methodshort{} ($B=3$) &                 \methodshort{} ($B=4$) \\
\midrule
SST-2 & 2   &               $93.5_{1.6}$ &  $\boldsymbol{94.1_{1.3}}$ &               $93.8_{1.1}$ &               $94.0_{1.1}$ \\
CR & 2   &  $\boldsymbol{93.9_{0.7}}$ &               $93.8_{0.7}$ &               $\boldsymbol{93.9_{0.7}}$ &               $92.9_{1.5}$ \\
RTE & 2   &               $58.3_{3.8}$ &  $\boldsymbol{59.4_{3.9}}$ &               $58.1_{3.7}$ &               $57.9_{4.3}$ \\
Subj & 2   &  $\boldsymbol{84.1_{7.7}}$ &               $81.9_{7.5}$ &               $79.1_{7.2}$ &               $77.7_{7.0}$ \\
CB & 3   &  $\boldsymbol{65.2_{8.0}}$ &               $59.9_{7.7}$ &               $61.2_{8.2}$ &               $56.8_{5.4}$ \\
AGNews & 4   &               $79.8_{3.6}$ &               $80.9_{2.4}$ &               $81.5_{2.1}$ &  $\boldsymbol{81.9_{1.9}}$ \\
SST-5 & 5   &               $45.5_{3.9}$ &               $46.3_{3.9}$ &  $\boldsymbol{47.4_{2.9}}$ &               $46.1_{2.8}$ \\
YELP & 5   &               $56.2_{3.8}$ &  $\boldsymbol{56.8_{3.4}}$ &               $56.3_{5.1}$ &               $54.8_{3.1}$ \\
TREC & 6   &               $87.0_{4.5}$ &               $88.8_{3.4}$ &               $89.4_{3.2}$ &  $\boldsymbol{89.7_{3.0}}$ \\
DBPedia & 14  &               $93.2_{3.0}$ &               $95.1_{2.3}$ &               $96.2_{1.5}$ &  $\boldsymbol{96.4_{1.3}}$ \\
NLU Scenario & 18  &               $81.9_{2.2}$ &               $83.4_{1.7}$ &               $84.2_{1.5}$ &  $\boldsymbol{84.6_{1.4}}$ \\
TREC Fine & 50  &               $60.5_{6.9}$ &               $65.2_{3.8}$ &  $\boldsymbol{68.8_{3.4}}$ &               $\boldsymbol{68.8_{3.2}}$ \\
NLU Intent & 68  &               $69.7_{3.3}$ &               $77.7_{2.1}$ &               $79.7_{1.9}$ &  $\boldsymbol{80.9_{1.9}}$ \\
BANKING77 & 77  &               $51.0_{3.4}$ &               $58.7_{3.3}$ &               $63.5_{2.7}$ &  $\boldsymbol{65.8_{2.5}}$ \\
CLINIC150 & 150 &               $67.3_{2.7}$ &               $74.4_{2.5}$ &               $75.4_{1.7}$ &  $\boldsymbol{78.1_{2.1}}$ \\
\bottomrule
\end{tabular}
    \caption{Results for different choices of $B$ for J1-Large model. The best result for each dataset is boldfaced.}
    \label{tab:large-diff-ks}
\end{table*}

\begin{table*}[t!]
\small
\centering
\begin{tabular}{lc llll}
\toprule
          Dataset &  \# Labels   &           ICL &                 \methodshort{} ($B=2$) &                 \methodshort{} ($B=3$) &                 \methodshort{} ($B=4$) \\
\midrule
SST-2 & 2   &  $95.2_{1.1}$ &               $95.4_{0.7}$ &  $\boldsymbol{95.6_{0.5}}$ &               $\boldsymbol{95.6_{0.3}}$ \\
CR & 2   &  $93.6_{0.8}$ &  $\boldsymbol{93.9_{0.9}}$ &               $93.8_{0.8}$ &               $\boldsymbol{93.9_{0.7}}$ \\
RTE & 2   &  $61.2_{5.1}$ &  $\boldsymbol{64.2_{2.7}}$ &               $62.2_{3.0}$ &               $62.4_{3.4}$ \\
Subj & 2   &  $93.0_{2.5}$ &               $94.6_{1.3}$ &               $95.3_{1.2}$ &  $\boldsymbol{95.7_{1.0}}$ \\
CB & 3   &  $75.0_{8.1}$ &               $74.7_{8.3}$ &  $\boldsymbol{75.7_{6.0}}$ &               $73.0_{5.6}$ \\
AGNews & 4   &  $81.4_{3.0}$ &               $82.1_{2.4}$ &  $\boldsymbol{82.7_{2.1}}$ &               $82.6_{2.0}$ \\
SST-5 & 5   &  $51.6_{3.4}$ &               $53.6_{2.9}$ &               $53.8_{2.2}$ &  $\boldsymbol{53.9_{2.0}}$ \\
YELP & 5   &  $66.2_{2.2}$ &  $\boldsymbol{66.6_{1.7}}$ &               $65.6_{2.0}$ &               $65.5_{1.9}$ \\
TREC & 6   &  $86.5_{3.8}$ &               $88.1_{4.0}$ &               $88.7_{3.4}$ &  $\boldsymbol{89.2_{4.5}}$ \\
DBPedia & 14  &  $92.5_{3.3}$ &               $95.8_{2.7}$ &               $97.3_{1.6}$ &  $\boldsymbol{97.9_{1.3}}$ \\
NLU Scenario & 18  &  $86.1_{2.1}$ &               $88.4_{1.4}$ &               $88.8_{1.1}$ &  $\boldsymbol{89.2_{1.2}}$ \\
TREC Fine & 50  &  $63.3_{6.0}$ &               $67.7_{4.3}$ &  $\boldsymbol{71.8_{4.6}}$ &               $71.2_{4.8}$ \\
NLU Intent & 68  &  $72.1_{3.1}$ &               $79.7_{2.4}$ &               $81.9_{1.6}$ &  $\boldsymbol{83.3_{1.6}}$ \\
BANKING77 & 77  &  $55.2_{3.3}$ &               $64.5_{3.1}$ &               $69.1_{2.2}$ &  $\boldsymbol{70.9_{2.8}}$ \\
CLINIC150 & 150 &  $68.9_{2.5}$ &               $76.5_{2.5}$ &               $78.6_{1.8}$ &  $\boldsymbol{80.2_{2.6}}$ \\
\bottomrule
\end{tabular}
    \caption{Results for different choices of $B$ for J1-Grande model. The best result for each dataset is boldfaced.}
    \label{tab:grande-diff-ks}
\end{table*}

\begin{table*}[t!]
\small
\centering
\begin{tabular}{lc lll}
\toprule
         Dataset &  \# Labels   &           ICL &                 \methodshort{} ($B=2$) &                 \methodshort{} ($B=3$) \\
\midrule
SST-2 & 2   &  $96.5_{1.4}$ &  $\boldsymbol{97.8_{0.9}}$ &               $97.0_{1.5}$ \\
CR & 2   &  $93.6_{1.5}$ &  $\boldsymbol{93.9_{1.0}}$ &               $93.1_{1.0}$ \\
RTE & 2   &  $63.9_{5.0}$ &               $65.2_{3.9}$ &  $\boldsymbol{66.0_{4.1}}$ \\
Subj & 2   &  $89.1_{5.3}$ &               $91.6_{3.0}$ &  $\boldsymbol{93.6_{2.1}}$ \\
CB & 3   &  $76.2_{4.3}$ &               $76.2_{7.1}$ &  $\boldsymbol{76.6_{3.5}}$ \\
AGNews & 4   &  $82.5_{3.8}$ &               $84.9_{1.7}$ &  $\boldsymbol{85.9_{1.7}}$ \\
SST-5 & 5   &  $55.4_{2.8}$ &  $\boldsymbol{55.6_{3.2}}$ &               $55.1_{3.9}$ \\
YELP & 5   &  $66.3_{4.1}$ &  $\boldsymbol{68.3_{2.5}}$ &               $65.4_{2.6}$ \\
TREC & 6   &  $87.1_{5.7}$ &               $89.1_{3.0}$ &  $\boldsymbol{90.4_{3.1}}$ \\
DBPedia & 14  &  $91.7_{4.4}$ &               $96.2_{2.6}$ &  $\boldsymbol{96.5_{2.3}}$ \\
NLU Scenario & 18  &  $85.4_{2.9}$ &               $87.1_{1.8}$ &  $\boldsymbol{87.8_{1.6}}$ \\
TREC Fine & 50  &  $71.4_{5.7}$ &               $77.5_{2.4}$ &  $\boldsymbol{78.7_{3.6}}$ \\
NLU Intent & 68  &  $74.3_{3.4}$ &               $80.3_{2.5}$ &  $\boldsymbol{81.6_{2.9}}$ \\
BANKING77 & 77  &  $55.3_{3.5}$ &               $65.9_{3.9}$ &  $\boldsymbol{70.9_{3.1}}$ \\
CLINIC150 & 150 &  $65.7_{5.0}$ &               $74.8_{4.2}$ &  $\boldsymbol{79.9_{2.1}}$ \\
\bottomrule
\end{tabular}
    \caption{Results for different choices of $B$ for J1-Jumbo model. The best result for each dataset is boldfaced. For computational considerations, we have only attempted to use $B=2$ and $B=3$.}
    \label{tab:Jumbo-diff-ks}
\end{table*}

\begin{table*}[t!]
     \small
    \centering
    \begin{tabular}{@{}p{0.08\textwidth}p{0.025\textwidth}p{0.05\textwidth}p{0.025\textwidth}p{0.38\textwidth}p{0.33\textwidth}@{}}
    \toprule
    \textbf{Dataset} & \textbf{$n_{max}$ J1} & \textbf{$n_{max}$ LLaMA} & \textbf{$n_{max}$ GPT2} & \textbf{Prompt Example} & \textbf{Labels}\\
    \midrule
    SST-2 & 68 & 48 & 27 & Review: , the power of this movie is undeniable . \newline Sentiment: positive  & [negative, positive] \\
    \midrule
    CR & 54 & 39 & 21 & premise: Review: in fact , i liked it so much after using it with my son who is now 2 years old , that i bought one for our new baby ' s room\newline Sentiment: positive &  [negative, positive] \\
    \midrule
     RTE & 17 & 10 & 5 & premise: The 10-day-old "test-tube" baby elephant born at Colchester Zoo has found a name, thanks to the internet!\newline hypothesis: baby elephant born\newline prediction: True & [True, False] \\
     \midrule
     Subj & 42 & 32 & 18 & Input: they follow him to las vegas , where he is ostensibly doing `` research `` for the next season , but is actually pursuing a dream to become a dancer in a vegas show .\newline Type: objective &[objective, subjective] \\
     \midrule
     CB & 19 & 10 & 5 & premise: Paula could not help herself. It was just the way she was. Others might say they hated her and mean it.\newline hypothesis: others hated Paula\newline prediction: true &[true, false, neither] \\
     \midrule
    AGNews & 30 & 20 & 11 & input: Citigroup faces regulatory probe The UK's Financial Services Authority launches a formal investigation into Citigroup's "unusual trading activity".\newline type: business & [world, sports, business, technology] \\
    \midrule
    SST-5 & 51 & 36 & 20 &Review: it 's just a little too self-satisfied .\newline Sentiment: okay & [terrible, bad, okay, good, great] \\
    \midrule
    YELP & 5 & 2 & 0 & review: Good modern atmosphere and delicious cupcakes.\newline stars: 3 & [1, 2, 3, 4, 5] \\
    \midrule
    TREC & 88 & 69 & 38 & Question: When was the first Barbie produced ?\newline Type: numeric & [abbreviation, entity, description, human, location, numeric] \\
    \midrule
    DBPedia & 21 & 14 & 7 &                                                            input: The Bstanu River is a tributary of the Râul Mare in Romania.\newline type: nature & [company, school, artist, athlete, politics, transportation, building, nature, village, animal, plant, album, film, book] \\
    \midrule
    NLU\newline Scenario & 112 & 80 & 43 &                                                     utterance: you have got the answer right.\newline scenario: general & [lists, weather, general, cooking, email, alarm, datetime, calendar, social, transport, iot, recommendation, takeaway, play, music, qa, news, audio] \\
    \midrule
    TREC Fine & 84 & 65 & 37 &                                                         Question: What dropped 1 , 313 feet in 1980 ?\newline Type: entity other & [abbreviation abbreviation, abbreviation expansion, entity animal, entity body, entity color, entity creation, entity currency, entity disease, entity event, entity food... \\
    \midrule
    NLU Intent & 101 & 80 & 43 &                                                       utterance: please read out the tasks from the list for today\newline intent: lists query & [alarm query, alarm remove, alarm set, audio volume down, audio volume mute, audio volume other, audio volume up, calendar query, calendar remove, calendar set... \\
    \midrule
    BANKING77 & 77 & 51 & 27 &                                                         query: Card payment didn't go through.\newline intent: declined card payment & [activate my card, age limit, apple pay or google pay, atm support, automatic top up, balance not updated after bank transfer, balance not updated after cheque or cash deposit... \\
    \midrule
     CLINIC150 & 101 & 72 & 39 &                                                                                         utterance: i would like to look up my credit score please\newline intent: credit score & [restaurant reviews, nutrition info, account blocked, oil change how, time, weather, redeem rewards, interest rate, gas type... \\
    \bottomrule
    \end{tabular}
    \caption{Table of classification datasets with their used prompts and the $n_{max}$ for both GPT2 and J1 tokenizers. For readability, we truncated the list of labels for some of the multi-label tasks.}
    \label{tab:prompts}
\end{table*}

\begin{table*}
\small
\centering
\begin{tabular}{p{3.0cm}p{12.0cm}}
\toprule
\textbf{Task} & \textbf{Prompt}\\
\midrule
Natural Questions (NQ)& Title: We Bought a Zoo\newline
Evidence: We Bought a Zoo We Bought a Zoo is a 2011 American family...  The film also stars Scarlett Johansson, Maggie Elizabeth Jones...\newline
Question: who is the little girl on we bought a zoo?\newline
Answer: Maggie Elizabeth Jones\newline
\vspace{3.9pt}\hspace{-0.1cm}
Title: Vaal River\newline
Evidence: ...The river flows west into the Grootdraai Dam near Standerton, Mpumalanga. On its course to the Vaal Dam in Vereeniging...\newline
Question: where does the vaal dam get its water from?\newline
Answer: Vaal River\newline
==\newline
Title: San Juan River (Colorado River tributary)\newline
Evidence: in the San Juan Mountains has often been diminished by warming winter temperatures..\newline
==\newline
Title: olorad\newline
Evidence: drained by the Colorado River. The South Park of Colorado is the region of the headwaters of the South Platte River...\newline
==\newline
Title: San Juan River (Colorado River tributary)\newline
Evidence: ...Colorado at the confluence of its East and West Forks. Both forks originate above elevations of in the eastern San Juan Mountains in the San Juan National Forest...\newline
==\newline
Question: where are the san juan mountains in new mexico?\newline
Answer: \\
\midrule
HotpotQA & Evidences:\newline
==\newline
The 2009 Singapore Grand Prix (formally
the 2009 Formula 1 SingTel Singapore
Grand Prix) was a Formula One motor race
held at the Marina Bay Street Circuit in
Singapore on 27 September 2009...\newline
==\newline
Catharina Felser (born October 2, 1982) is
a German race car driver born in Siegburg...\newline
==\newline
...Sergio Pérez, the only other Mexican to
finish on the podium, currently races with
Sahara Force India F1 Team .\newline
==\newline
Sergio Pérez Mendoza ( ; born 26 January
1990) also known as "Checo" Pérez, is a
Mexican racing driver, currently driving for
Force India.\newline
==\newline
Question: Which other Mexican Formula
One race car driver has held the podium
besides the Force India driver born in 1990?\newline
Answer: \\
\bottomrule
\end{tabular}
\vspace{-0.2cm}
\caption{Prompt formats for Natural Questions (NQ) and HotpotQA. The prompts were manually shortened for readability.}
\label{tab:nq-hotpotqa-format}
\end{table*}

\end{document}